\documentclass[10pt,twocolumn,letterpaper]{article}

\usepackage{booktabs}       
\usepackage{amsfonts}       
\usepackage{nicefrac}       
\usepackage{microtype}      
\usepackage{xcolor}         
\usepackage{multirow}
\usepackage{amsmath}
\usepackage{algorithm}
\usepackage{xcolor}         
\usepackage{booktabs} 
\usepackage{multirow} 
\usepackage{makecell}
\usepackage{pifont} 
\usepackage{circledsteps}
\usepackage{makecell}
\usepackage{tabularx}
\usepackage[accsupp]{axessibility}  
\usepackage{graphicx}
\usepackage{amssymb}
\usepackage{subcaption}
\usepackage{bbding}
\usepackage[accsupp]{axessibility} %
\usepackage{algpseudocode}
\usepackage{algorithm}
\usepackage{nicefrac}
\usepackage{booktabs}
\usepackage{multirow}
\usepackage{multicol}
\usepackage{makecell}
\usepackage{colortbl}
\usepackage{arydshln}
\usepackage{subcaption}
\usepackage{amssymb}
\usepackage{pifont}
\usepackage{setspace}
\usepackage{tcolorbox}
\usepackage{wrapfig}
\usepackage{multirow}
\usepackage[normalem]{ulem}
\useunder{\uline}{\ul}{}

\definecolor{myred}{RGB}{200,50,50}

\newcommand{\eg}{\emph{e.g.}{}}

\newcommand{\cf}{\emph{cf. }{}}

\definecolor{mycustomcolor}{HTML}{DDEEFF} 

\definecolor{crosscolor}{rgb}{0.969,0.580,0.114} %
\definecolor{checkcolor}{rgb}{0.485,0.640,0.204} %

\usepackage{array} 
\newcolumntype{x}[1]{>{\centering\arraybackslash}p{#1pt}}

\newlength\savewidth

\usepackage[pagenumbers]{cvpr} 

\definecolor{cvprblue}{rgb}{0.21,0.49,0.74}
\usepackage[pagebackref,breaklinks,colorlinks,allcolors=cvprblue]{hyperref}

\def\paperID{871} 
\def\confName{CVPR}
\def\confYear{2026}

\title{Latent Visual States for Efficient Multimodal Reasoning}

\author{
{\bf Xiuwei Chen$^1$}, ~
{\bf Wentao Hu$^2$,} ~
{\bf Yongxin Wang$^3$,} ~
{\bf Zisheng Chen$^1$,} ~
\\
{\bf Likui Zhang$^1$,} ~
{\bf Kun Xiang$^1$,} ~
{\bf Jianhua Han$^5$,}
{\bf Hui-Ling Zhen$^4$,} ~ 
\\
{\bf Jingyuan Zou$^4$,} ~
{\bf Hang Xu$^5$,} ~
{\bf Xiaodan Liang$^{1}$}\thanks{Corresponding author.}
\\
\\
$^{1}$Sun Yat-sen University 
$^{2}$The Hong Kong Polytechnic University \\
$^{3}$MBZUAI
$^{4}$Huawei Noah's Ark Lab 
$^{5}$Yinwang Intelligent Technology Co. Ltd.
}

\begin{document}
\maketitle
\begin{abstract}
The integration of visual evidence has significantly enhanced the capabilities of large multimodal models. However, this integration predominantly relies on generating discrete outputs (\eg, code or box coordinates) to invoke external tools, a process that introduces rigid dependencies and substantial latency.
To overcome these limitations, we propose {EVA} (Lat{\bf E}nt {\bf V}isual St{\bf A}tes), a novel framework that natively generates continuous latent visual representations.
These internal representations manifest as an adaptive sequence of \texttt{Latent\_slot} tokens, serving as intermediate visual thoughts during the reasoning process.
These \texttt{Latent\_slot} tokens are then trained end-to-end with the discrete text tokens.
This co-optimization, notably, causes extreme policy deviation in the 'transition window' following the \texttt{Latent\_slot} tokens. 
We develop D-GSPO (Decouple-GSPO) to target this root cause by decoupling the optimization of latent and discrete components.
To support SFT, we construct EVA-230K, a high-quality text-image interleaved CoT dataset encompassing a diverse range of real-world scenes, documents, charts and OCR tasks.
Extensive experiments across multiple benchmarks confirm that EVA achieves significant performance gains while enhancing inference efficiency.
\end{abstract}    
\section{Introduction}
\label{sec:intro}

\begin{figure*}[t]
	\centering
        \includegraphics[width=1.0\textwidth]{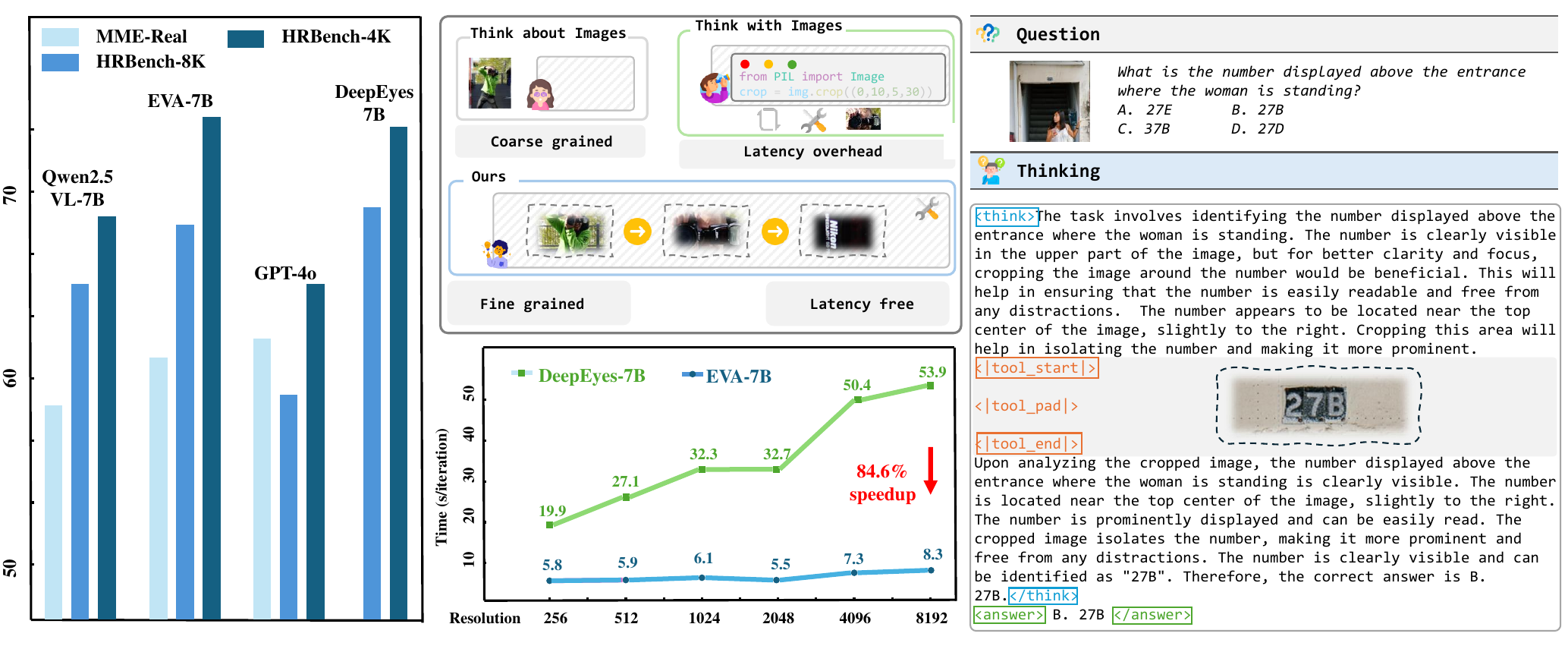}
        \caption{{\bf Performance and Paradigm Comparison of the Proposed EVA Framework.}
        (Left) Comparative results with alternative methods demonstrate that EVA achieves superior overall performance. (Middle) Contrasting different paradigms, EVA internalizes the explicit external tool invocation into the model's intrinsic capabilities; this mechanism is validated by the significantly lower latency observed when compared to methods utilizing external tools. (Right) The generated outputs from the EVA framework are shown, visually confirming the efficacy.}
	\label{fig:fig_compare}
	\vspace{-2.5mm}
\end{figure*}

{Multimodal large language models (MLLMs) have achieved remarkable progress in visual reasoning~\cite{zhu2023minigpt, bai2025qwen2, hurst2024gpt, o3}, yet challenges remain for complex, fine-grained perception~\cite{huang2025vision, chen2025mint, zhang2025chain}. This motivates a transition toward more active and reasoning-driven visual understanding.}
{Early efforts to enhance reasoning in MLLMs primarily extended the Chain-of-Thought (CoT) paradigm to the visual domain, enabling models to \textit{think about images} by producing step-by-step textual rationales before answering~\cite{LLaVACoT,R1VL,R1Onevision}. While visual CoT improves transparency and multi-step planning, purely textual traces are often insufficient for fine-grained perception: many problems require manipulating the visual input itself (\eg, cropping, or re-focusing) to reveal details that are otherwise inaccessible to the model (\cf, Figure~\ref{fig:fig_compare}). This realization has spurred a second wave of approaches that encourage models to \textit{think with images} by invoking external visual tools during inference~\cite{feng2025retool, zhang2025thyme, lai2025mini, zheng2025deepeyes, zhang2025chain, hu2024visual}. MLLMs sequentially issue discrete commands, such as bounding boxes, programmatic crop calls, or detector triggers, to obtain refined visual inputs that guide downstream reasoning.}

{Although external visual tool use has demonstrated compelling accuracy gains, it also introduces two fundamental limitations (\cf, Figure~\ref{fig:fig_compare}). \textbf{Representational decoupling} arises because discrete tool calls sit outside the model’s internal computation: the model must halt its reasoning, emit symbolic commands, wait for non-differentiable utilities to execute, and then re-encode the results. This stop–and–go workflow prevents end-to-end optimization of the interplay between how visual evidence is acquired and how that evidence is used for reasoning. In parallel, \textbf{operational inefficiency} accumulates from multi-step processes, repeated visual re-encoding, and rigid dependencies on external runtimes, costs that become especially prohibitive for high-resolution inputs or multi-turn visual exploration.}

To overcome these limitations, we introduce Lat{\bf E}nt {\bf V}isual St{\bf A}tes (EVA), a framework that leverages an adaptive sequence of continuous internal visual representations to serve as intermediate visual thoughts during reasoning.
Fundamentally, EVA is designed to generate the continuous latent visual space that acts as a proxy for the target visual features, thereby eliminating the dependency on external utilities.
Rather than aligning with all visual features, we employ a saliency-weighting scheme to identify task-specific semantic regions, thereby mitigating interference from irrelevant background context.
Furthermore, the framework is designed to be adaptive, allowing it to generate a variable number of Latent\_slots based on the task's context. To enable this adaptive capability, we construct a large-scale dataset, named EVA-SFT-230K. 
Each sample is carefully designed to contain a random and context-dependent number of Latent\_slots.

Building upon the EVA framework and dataset, we devise a multi-stage training strategy to progressively enhance the model's reasoning abilities. This process unfolds in three distinct phases: 1) learning to generate Latent\_slots (SFT), 2) learning to utilize them (SFT), and 3) learning to explore (GSPO). During the RL training stage, we empirically observe a significant policy deviation within the "transition window" that follows the Latent\_slots (\eg, the five tokens immediately following the Latent\_slot), diverging sharply from other token positions, which leads to training instability in the form of exploding gradients.
To address this instability, we introduce the Decouple-GSPO (D-GSPO) algorithm, which computes the policy objective solely on discrete text probabilities (while still allowing gradients to flow back through the latent Latent\_slot).
and simultaneously applying a stronger, localized KL constraint specifically to the unstable 'transition window' tokens.
This targeted separation effectively preserves the model's exploratory capacity without compromising stability.

We validate the effectiveness of our complete approach on multiple benchmarks, achieving consistent improvements. Furthermore, our framework demonstrates superior inference efficiency across various resolutions when compared to conventional tool-invocation methods (\cf, Figure~\ref{fig:fig_compare} (Middle)).
Our contributions are summarized as follows:

    {$\bullet$ We propose EVA, a novel framework that natively generates the adaptive sequence of continuous latent visual representations (Latent\_slots), enabling end-to-end optimization of reasoning and visual understanding within the model.}

    $\bullet$ We develop a three-stage training strategy for progressively enhancing the model's tool capabilities. To address the high policy deviation occurring in the GSPO "transition window," we introduce the D-GSPO algorithm, which effectively mitigates the instability arising from the joint optimization of continuous and discrete representations.

    $\bullet$ We validate EVA on multiple benchmarks, demonstrating that our method not only achieves consistent effectiveness but also exhibits significantly faster inference efficiency compared to conventional tool-invocation methods.
\section{Related Work}
\label{sec:rel}

\subsection{Latent Reasoning}
Conventional Chain-of-Thought (CoT) reasoning is typically based on human-interpretable language forms. In contrast, the Coconut~\cite{hao2024training} introduced the concept of continuous chains of thought to replace traditional linguistic formats, arguing that their latent representations were richer than those of discrete text. However, initial experiments in the original Coconut paper did not outperform models relying on traditional language-based reasoning. This limitation subsequently motivated a series of follow-up studies and optimizations, including efforts such as LatentSEEK \cite{li2025seek}, and \cite{li2025implicit}. 
More recently, similar continuous chain-of-thought paradigms have been introduced in domains such as maze solving and mathematics, where models often represent visual or intermediate information using latent tokens.
For example, 
AURORA~\cite{bigverdi2025perception} enhances the reasoning capabilities of Multimodal Language Models (MLMs) on fine-grained visual perception tasks (\eg, counting and depth estimation) by leveraging a VQVAE to transform intermediate visual information, such as depth maps and bounding boxes, into Perception Tokens.
Mirage~\cite{mirage} enhances the model’s spatial reasoning ability by incorporating visual latent tokens in addition to textual inputs. 
MINT-CoT~\cite{chen2025mint} introduces a visual-encoder-guided supervision mechanism, aligning visual latent tokens with semantically meaningful visual features. This allows the model to generate reasoning-relevant latent representations during the reasoning process, leading to notable improvements in mathematical reasoning tasks.
LVR ~\cite{lvr} further extends this line of work by proposing GRPO$_{latent}$, a variant that can be seamlessly integrated into any latent reasoning model, achieving competitive results on three benchmarks.
In contrast to the aforementioned methods, our approach internalizes the explicit external tool invocation directly into the model's core mechanism but operates within a more general visual domain. Furthermore, unlike prior approaches, our framework is inherently adaptive, focusing on dynamic adjustment during inference, a process that more closely aligns with human visual perception.
More related work can be found in {\color{blue} \bf APPENDIX G}.

\section{Method}
\label{sec:met}

\subsection{Overview}
\begin{figure*}[t]
	\centering
        \includegraphics[width=0.98\textwidth]{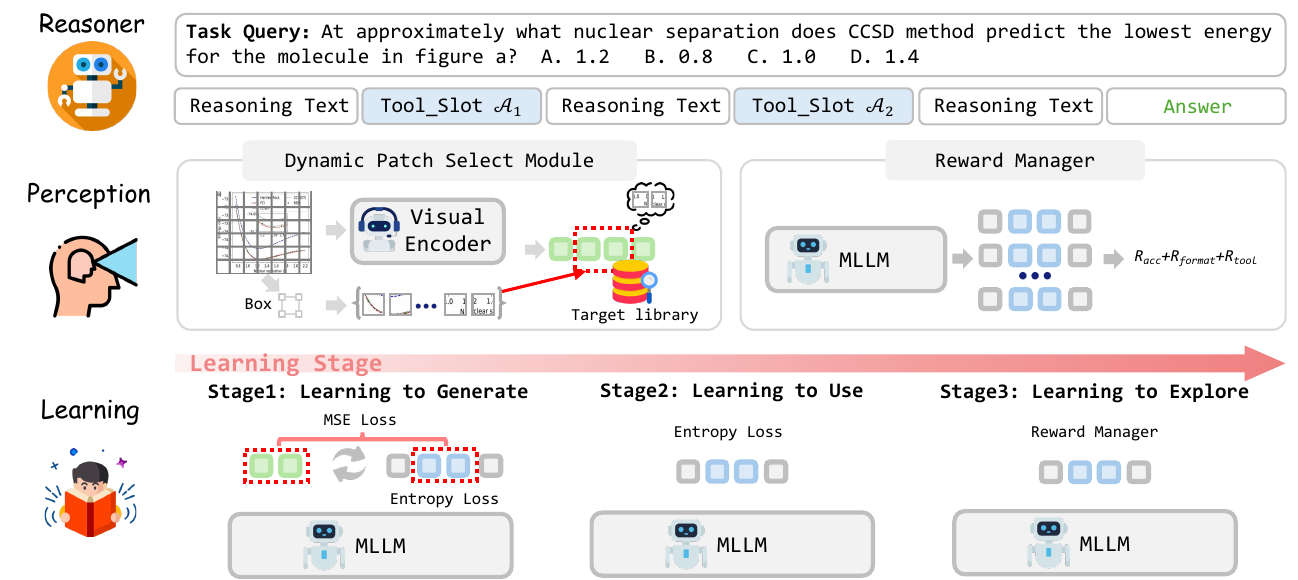}
        \caption{\textbf{Overall of the EVA framework}. }
	\label{fig:fig_main}
\end{figure*}
{\bf Overall EVA Pipeline} We illustrate the overall pipeline in Fig \ref{fig:fig_main}. Given an input query $Q$ and image $I$, our model, consisting of an LLM backbone $\pi_{\theta}$ and a visual encoder $E_v$, autoregressively generates a sequence of "Thought-Visual" pairs: $(T_1, \mathcal{A}_1), (T_2, \mathcal{A}_2), ..., (T_{N-1}, \mathcal{A}_{N-1}), (T_{N})$. Here, $T_i$ is a textual reasoning trace, and $\mathcal{A}_i$ is the 'Latent\_slot', a latent embedding representing visual thoughts. Crucially, the sequence length $N$ is not fixed; it is adaptively determined based on the model’s internal assessment of task difficulty. This process concludes when the model gathers sufficient evidence, terminating the sequence and producing the final answer.

\begin{itemize} 
\item Thought $T_i$: This component enables an iterative refinement process by synthesizing the input query $Q$, image $I$, and previously generated responses (\eg, $T_{i-1}, \mathcal{A}_{i-1}$).
This progressive mechanism, simulating human-like reflection, builds upon prior insights and visual abstractions to guide the generation of the subsequent visual Latent\_slot ($\mathcal{A}_i$).

\item Latent\_slot $\mathcal{A}_i$: Guided by the preceding thought $T_i$, this component represents the visual thought deemed relevant at the current reasoning stage. For example, when the query involves identifying a tiger, $\mathcal{A}_i$ corresponds to a latent embedding of the tiger region. This abstraction is structurally demarcated within the model's output sequence by special tokens (\eg, $\operatorname{<|latent\_start|> ... <|latent\_end|>}$). This encoding allows for seamless integration back into the model’s internal state, thereby informing the subsequent reasoning step ($T_{i+1}$).
We allocate a corresponding number of \texttt{Latent\_slot} placeholders based on the quantity of tool images present in each sample. To guarantee accurate alignment during the matching process with the ground truth, we use order constraints. 
The resulting features, preserved according to their original spatial relationships, are then compressed into a default n-dimensional \texttt{Latent\_slot}. This operation strictly ensures that the dimension of these slots remains consistent with that of the ground-truth regions. Finally, during the inference phase, we set a fixed maximum slot length. 
Furthermore, in the Reinforcement Learning (RL) phase, we discard a small number of samples generated during the rollout process where the quantity of predicted \texttt{Latent\_slot} tokens does not match the number of reserved positions.
Further details can be found in the {\color{blue} \bf APPENDIX C.5}.

\item Adaptive ability: To equip the model with the ability to dynamically adjust the number of visual embeddings (\eg, the count of $\mathcal{A}_i$ slots) according to task complexity, we construct a supervised fine tuning dataset. This dataset explicitly covers a spectrum of reasoning depths, from cases requiring zero $\mathcal{A}_i$ to those demanding up to three iterative steps. The model is then fine tuned on this curated dataset, enabling it to learn not just how to generate a $(T_i, \mathcal{A}_i)$ pair, but when and how many times to invoke this entire process in a data driven and task adaptive manner. 

\end{itemize}

\noindent {\bf Three-phase Training}

\begin{itemize}
\item Learning to Generate: We first fine-tune the model on our curated dataset. This initial stage endows the model with the foundational ability to adaptively generate visual abstractions and produce robust reasoning patterns.

\item Learning to Use: We observe that direct reinforcement learning causes catastrophic forgetting of this generative capability. This intermediate reinforcement phase is therefore introduced to explicitly preserve and enhance the model’s ability to leverage these learned abstractions.

\item Learning to Explore: Finally, this stabilized policy is optimized using D-GSPO.
\end{itemize}

\subsection{Learning to Generate \& Use}
\label{sec:3.2}

\begin{algorithm}[t]
   \caption{{EVA} Algorithm}
   \label{alg:training_process}
    \textbf{Input:} Cold-Start Dataset $\mathcal{D}_{sft}$, GSPO Dataset $\mathcal{D}_{GSPO}$, Base Model $\pi_{\theta}$
     
    \textbf{Output:} Improved Model $\pi_{\theta, \text{GSPO}}$
     
   \begin{algorithmic}[1]

    \State \textcolor{gray}{$\triangleright$  \textbf{Learning to Generate \& Use}} \Comment{see §\ref{sec:3.2}}
    \State $\pi_{\theta, SFT1}  \leftarrow$ update $\pi_{\theta}$ with SFT on $\mathcal{D}_{sft}$ \Comment{Using Eq \ref{eq:l_total}}
    

    \State $\pi_{\theta, SFT2}  \leftarrow$ update $\pi_{\theta, SFT1}$ with SFT on $\mathcal{D}_{sft}$ \Comment{Using Eq \ref{eq:l_ce}}

    \State \textcolor{gray}{$\triangleright$  \textbf{Learning to Explore}} \Comment{see §\ref{sec:3.3}}
    \For{each batch $B$ in $D_{GSPO}$}
    \State $T_i, \mathcal{A}_i \leftarrow \pi_{\theta, SFT2}$(B)
    \State $r_i \leftarrow \operatorname{Tool\_Reward}(T_i, \mathcal{A}_i)$ 
    \State $\pi_{\theta, \text{GSPO}} \leftarrow$ D-GSPO($\pi_{\theta, SFT2},T_i,\mathcal{A}_i,r_i$) \Comment{Using Eq \ref{eq:l_gspo}}
    \EndFor

   \end{algorithmic}

    
    



\end{algorithm}

The primary objective of the EVA framework is to generate latent visual representations instead of invoking external visual tools. EVA is realized through a series of adaptive Latent\_slots, which reframe discrete tool operations as a learned, intrinsic capability. The successful implementation of this mechanism hinges on two key components. We first detail the optimization objective for the Latent\_slot, which we formulate as the Dynamic Patch Selection Module. We then describe the Cold-Start Data Construction process required to bootstrap this capability.

\begin{figure}[t]
	\centering
        \includegraphics[width=0.5\textwidth]{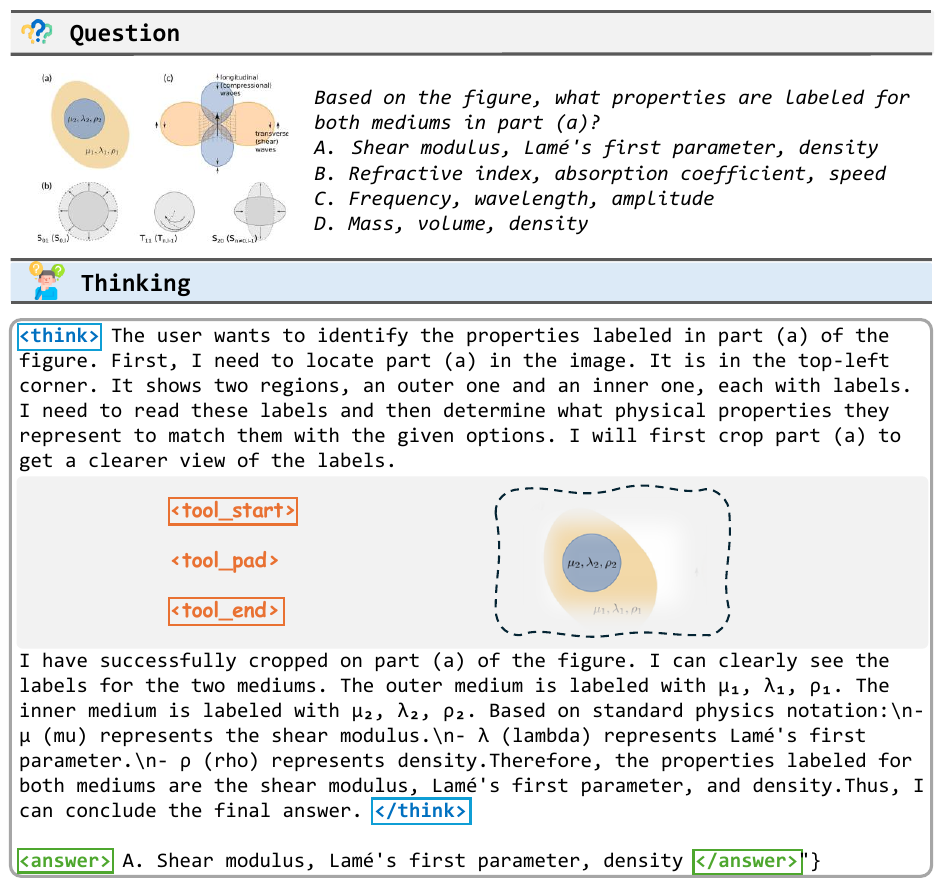}
        \caption{\textbf{Qualitative Examples of Our Dataset.} Each sample illustrates the model's reasoning process (think), the corresponding localized visual regions, and the final answer.}
	\label{fig:fig_data}
\end{figure}

\noindent {\bf Dynamic Patch Select Module.} 
The optimization target for the Latent\_slot $\mathcal{A}_i$ is to generate an embedding that serves as a proxy for an target visual images. This target representation is formulated from the target visual regions identified in the annotated reasoning trace, using their underlying patch-level features. For each training sample $(Q, I)$, we first obtain its complete set of patch-level embeddings $\{e_1, e_2, \dots, e_K\}$. 
Concurrently, we adopt a task-specific selection strategy: for the \textit{cropping} task, we utilize the ground-truth bounding box to identify the subset of patch features $\{e_1, \dots, e_M\}$ whose centers are enclosed by the specified regions. For other tasks such as \textit{rotation} and \textit{drawing}, we retain the original features of the target image to preserve its global context.
Specifically, we compute an affinity map between the visual features and the task-specific semantic embeddings extracted from the input query, and preserve the highly-scored features as a set of target features for subsequent reasoning.
In accordance with our adaptive multi-slot design, this entire selection and projection process is applied independently for each ground-truth in the sample. The complete set of resulting target vectors $\{r_{\text{target}}^1, \dots, r_{\text{target}}^{N-1}\}$ is aggregated to form the target library  $\mathcal{R}_{target}$. This library serves as the ground-truth reference, providing the essential visual evidence that the Latent\_slot embeddings $\mathcal{A}$ are trained to predict, with the set of generated $\mathcal{A}_i$ embeddings being optimized to match the set of vectors within this library.
Subsequently, we optimize each generated Latent\_slot $\mathcal{A}_i$ against the target library $\mathcal{R}$ using the Mean Squared Error (MSE) loss:
\begin{equation}
{\mathcal{L}_{MSE}} = \frac{1}{{|\mathcal{A}|}}\sum\limits_{i = 1}^{|\mathcal{A}|} {||{\mathcal{A}_i} - {R_i}|{|^2}}
\label{eq:l_mse}
\end{equation}
Additionally, for the Thought ($T$), we utilize the Cross-Entropy (CE) loss, as shown in the following equation:
\begin{equation}
\mathcal{L}_{CE} = \frac{1}{|T|} \sum_{t=1}^{|T|} -\log p(\hat{y}_t \mid I, {Q}, y_{<t},\mathcal{A}_{<t}) 
\label{eq:l_ce}
\end{equation}
where $\hat{y}_t$ denotes the ground-truth token. The overall training loss is a weighted sum of the two components:
\begin{equation}
\mathcal{L}_{total} =  \mathcal{L}_{\mathrm{CE}} + \lambda_{\mathrm{MSE}} \cdot \mathcal{L}_{\mathrm{MSE}},
\label{eq:l_total}
\end{equation}

\noindent {\bf SFT Data Construction.} 
To build a dataset that supports our adaptive multi-slot design, we introduce the EVA-SFT Dataset, curating and processing data from several open-source benchmarks, including DeepEyes~\cite{zheng2025deepeyes}, ReFocus~\cite{fu2025refocus}, Thyme~\cite{zhang2025thyme}, and Visual-CoT~\cite{shao2024visual}. 
Initially, we collect a data subset containing between one and six tool images per sample and their corresponding reasoning chains. For selected samples from Visual-CoT and Refocus, we employ the Qwen3-VL-32B model guided by prompts and utilizing the retained tool images to generate detailed reasoning chains. To ensure that these samples contain multiple rounds of tool invocations, we expand the generated data using new prompts alongside a broader range of tool images that includes the original ground truth tool images. This expanded visual context encourages the model to conduct deeper analysis and consequently produce more tool invocations. Finally, we aggregate all the collected data and process all reasoning chains once more using the Qwen model. This final step specifically revises segments containing explicit code execution into an latent tool invocation format as illustrated in Figure~\ref{fig:fig_data}.
For the subsequent RL phase, we constructed our EVA-RL Dataset by sampling 4K samples from the DeepEyes~\cite{zheng2025deepeyes} and ThinkLite~\cite{wang2025sota} datasets.
Further details can be found in the {\color{blue} \bf APPENDIX A.1}.

\noindent{\bf Learning to Use} To ensure the model can reliably utilize these Latent\_slots during the RL phase, we compute the policy objective solely on the textual components; the Latent\_slots themselves, autoregressively generated by the policy, are thus implicitly optimized based on their ability to guide this subsequent text toward a higher reward.

\begin{table*}[]
\centering
\caption{{\bf Performance Comparison on Perception Tasks.} For all open-source models, the best performance for each metric is {\bf bolded}, and the second best is {\ul underlined}. {\color{brown} Gold-colored} font indicates improvement over the baseline Qwen2.5-VL-7B.
The result ($^{\dagger}$, $^{\ddagger}$) is collected from \cite{lvr} and \cite{lai2025mini}, respectively.
}
\label{tab:tab_main_results}
\scalebox{0.93}{

\begin{tabular}{lc
>{\columncolor[HTML]{DDEEFF}}c cc|ccc}
\toprule
\textbf{Benchmark}           & \multicolumn{1}{c}{\textbf{Split}} & \textbf{\begin{tabular}[c]{@{}c@{}}EVA-VL 7B\end{tabular}}                & \textbf{\begin{tabular}[c]{@{}c@{}}Qwen2.5-VL 7B\end{tabular}} & \textbf{\begin{tabular}[c]{@{}c@{}}Qwen2.5-VL 32B\end{tabular}} & \textbf{GPT-4o}             & \multicolumn{1}{c}{\textbf{\begin{tabular}[c]{@{}c@{}}LVR 7B$^{\dagger}$\end{tabular}}} & \textbf{\begin{tabular}[c]{@{}c@{}}DeepEyes 7B$^{\ddagger}$\end{tabular}} \\ \midrule
                             & FSP                                & \textbf{90.5\tiny{\color{brown}{+5.3}}} & 85.2                                                             & {\ul 87.5}                                                        & {\color[HTML]{656565} 66.8} & -                                                                             & -                                                              \\
                             & FCP                                & {\ul 56.8\tiny{\color{brown}{+4.6}}}    & 52.2                                                             & \textbf{59.3}                                                     & {\color[HTML]{656565} 63.3} & -                                                                             & -                                                              \\
\multirow{-3}{*}{HRbench-4K} & Overall                            & \textbf{73.7\tiny{\color{brown}{+4.9}}} & 68.8                                                             & {\ul 73.4}                                                        & {\color[HTML]{656565} 65.0} & -                                                                             & 73.2                                                           \\ \cline{2-8} 
                             & FSP                                & \textbf{83.3\tiny{\color{brown}{+4.5}}} & 78.8                                                             & {\ul 82.3}                                                        & {\color[HTML]{656565} 60.8} & -                                                                             & -                                                              \\
                             & FCP                                & {\ul 53.5\tiny{\color{brown}{+1.7}}}    & 51.8                                                             & \textbf{58.5}                                                     & {\color[HTML]{656565} 58.5} & -                                                                             & -                                                              \\
\multirow{-3}{*}{HRBench-8K} & Overall                            & 68.4\tiny{\color{brown}{+3.1}}          & 65.3                                                             & \textbf{70.4}                                                     & {\color[HTML]{656565} 59.6} & -                                                                             & {\ul 69.5}                                                     \\ \cline{2-8} 
                             & Perception                         & \textbf{63.9\tiny{\color{brown}{+3.3}}} & 60.6                                                             & {\ul 63.8}                                                        & {\color[HTML]{656565} 64.9} & \multicolumn{1}{c}{-}                                                         & -                                                              \\
                             & Reasoning                          & \textbf{41.8\tiny{\color{brown}{+3.2}}} & 38.6                                                             & {\ul 40.4}                                                        & {\color[HTML]{656565} 47.3} & \multicolumn{1}{c}{-}                                                         & -                                                              \\
\multirow{-3}{*}{MME-Real}      & Overall                            & \textbf{61.3\tiny{\color{brown}{+3.0}}}                         & 58.3                                                             & {\ul 61.0}                                                        & {\color[HTML]{656565} 62.8} & \multicolumn{1}{c}{-}                                                         & -                                                              \\ \cline{2-8} 
                                & Perception                         & \textbf{53.3\tiny{\color{brown}{+4.5}}}                         & 48.8                                                             & {\ul 50.6}                                                        & {\color[HTML]{656565} 54.4} & \multicolumn{1}{c}{-}                                                         & -                                                              \\
                                & Reasoning                          & \textbf{44.4\tiny{\color{brown}{+6.7}}}                         & 37.7                                                             & {\ul 39.3}                                                        & {\color[HTML]{656565} 48.3} & \multicolumn{1}{c}{-}                                                         & -                                                              \\
\multirow{-3}{*}{MME-Real-Lite} & Overall                            & \textbf{49.8\tiny{\color{brown}{+5.7}}}                         & 44.1                                                             & {\ul 46.2}                                                        & {\color[HTML]{656565} 52.0} & \multicolumn{1}{c}{-}                                                         & -                                                              \\ \cline{2-8} 
                                & Attribute                          & {\ul 80.0\tiny{\color{brown}{+1.8}}}                            & 78.2                                                             & 77.4                                                              & {\color[HTML]{656565} 72.2} & \textbf{84.4}                                                                 & -                                                              \\
                                & Spatial                            & {\ul 80.3\tiny{\color{brown}{+6.7}}}                            & 73.6                                                             & \textbf{86.8}                                                     & {\color[HTML]{656565} 60.5} & 77.6                                                                          & -                                                              \\
\multirow{-3}{*}{V*}            & Overall                            & 80.2\tiny{\color{brown}{+3.8}}                                  & 76.4                                                             & 81.2                                                              & {\color[HTML]{656565} 67.5} & {\ul 81.7}                                                                    & \textbf{83.3}                                                  \\ 
\bottomrule
\end{tabular}
}
\end{table*}


\subsection{Learning to Explore}
\label{sec:3.3}
In our initial reinforcement learning (GSPO) optimization, we attempted to jointly update both the discrete thought tokens ($T_i$) and the continuous Latent\_slot embeddings ($\mathcal{A}_i$) in an end-to-end manner. This approach, however, proved highly unstable and frequently led to gradient explosions. By visualizing the policy divergence, we observed that the "transition window" tokens immediately following each continuous slot $\mathcal{A}_i$ exhibited drastic distributional shifts. This indicates a critical form of catastrophic forgetting: the model loses its SFT-acquired ability to properly utilize the generated visual embedding ($\mathcal{A}_i$) to guide the subsequent reasoning step ($T_{i+1}$). This instability and the corresponding empirical observations, further detailed in the {\color{blue} \bf APPENDIX B}, motivate our D-GSPO algorithm.

To mitigate the aforementioned instability, we introduce Decouple-GSPO (D-GSPO). This framework optimizes the policy to maximize reward by decoupling the optimization of the discrete and continuous components.
For optimization, the policy's discrete tokens ($T_i$) are divided into two distinct parts: anchor regions (Latent\_slot $\mathcal{A}_i$) and explore regions ($T_i$ except anchor regions). 
Furthermore, a stronger KL loss is applied specifically to the transition window tokens.
We allow gradients to flow through the Latent\_slot ($\mathcal{A}_i$) tokens.
This overall decoupling prevents the policy gradient from destabilizing the SFT-learned visual representations.
The objective of the D-GSPO algorithm is to optimize the policy $\pi_\theta$ by maximizing the expected reward. This expectation is taken over inputs $x \sim D$ and the hybrid outputs $(T, \mathcal{A}) \sim \pi_\theta(\cdot|x)$, where $T$ and $\mathcal{A}$ are the sampled discrete tokens and continuous representations, respectively.
\begin{equation}
\begin{array}{l}
{\nabla _\theta }{J}(\theta ) = {\mathbb{E}_{x~D,\{ ({T_i},{\mathcal{A}_i})\} _{i = 1}^G~{\pi _\theta }( \cdot |x)}}[\frac{1}{G}\sum\limits_{i = 1}^G {\min ({s_i}(\theta ){{\hat {\rm A}}_i}} ) \\ - \beta {\mathbb{D}_{KL}}({\pi _\theta }||{\pi _{ref}})]
\end{array} 
\label{eq:l_gspo}
\end{equation}
To acquire a stable, low-variance advantage signal for discrete and continuous reasoning, we generate $g$ rollouts per input. We then normalize the reward for each response relative to its peers within the group.
\begin{equation}
{\hat A_i} = \frac{{{r_i} - \text{mean}([{r_1},{r_2}, \cdots ,{r_G}])}}{{\text{std}([{r_1},{r_2}, \cdots ,{r_G}])}}
\label{eq:l_ada}
\end{equation}
The other key component required for our optimization objective is the importance sampling ratio, $s_i(\theta)$. We define this ratio based on the sequence likelihood under the new and old policies, as shown in Equation~\ref{eq:l_ratio}:
\begin{equation}
\begin{array}{l}
\begin{gathered}
s_i(\theta) = \exp ( \frac{1}{\sum\nolimits M_{i,t}} \sum_{t=1}^{|y_i|} M_{i,t}   \log  \frac{\pi_\theta(y_{i,t} \mid x_i, T_{i<t}, \mathcal{A}_i)}{\pi_{\theta_{\text{old}}}(y_{i,t} \mid x_i, T_{i<t}, \mathcal{A}_i)}) \\
M_{i,t} = \mathbb{I}\left( z_{i,t} = \text{explore} \right)
\end{gathered}
\end{array} 
\label{eq:l_ratio}
\end{equation}
where $z_{i,t}$ denote the region assignment for the $t$-th token of the $i$-th response. 
To optimize the policy, we designed a composite reward function based on accuracy and format. The primary component is a binary accuracy reward (1 if the $\operatorname{<answer>}$ matches the ground-truth, 0 otherwise). This is supplemented by a format reward, which is positive only if the output strictly adheres to the $\operatorname{<think>...</think><answer>...</answer>}$ template. 
\section{Experiment}
\label{sec:exp}

\subsection{Experimental Settings}

\noindent {\bf Implementation Details}
In our experiments, we adopt a three-stage training strategy, where both the first and second stages utilize supervised fine-tuning (SFT) data with one epoch each. We initialize our framework from the open-source multimodal large language model Qwen2.5-VL-7B~\cite{bai2025qwen2} as the baseline. The learning rates for the first and second SFT stages are set to \(1 \times 10^{-5}\) and \(1 \times 10^{-6}\), respectively, while all other hyperparameters remain unchanged across these stages. Specifically, we employ a warmup ratio of 0.03, a weight decay of 0.1, and a cosine learning rate scheduler. The model obtained at the end of the second SFT stage is then used to initialize the subsequent reinforcement learning (RL) phase. The hyperparameter $\lambda_{\mathrm{MSE}}$ is set to 1.

For the GSPO~\cite{zheng2025group} stage, we build upon the VLM-R1 framework as the foundation of our implementation. In this phase, we train for one epoch with a learning rate of \(1 \times 10^{-5}\), a maximum completion length of 2048 tokens, and a total batch size of 128. We set the KL divergence coefficient to 0.04 and generate eight responses per input for reward computation. 
All experiments are conducted using NVIDIA H800 GPUs.
For a fair and rigorous efficiency comparison, all experiments are conducted on a single NVIDIA H800 GPU to maintain a consistent hardware environment. Under this unified hardware context, we ensure that no other concurrent processes are running to eliminate external interference during the latency measurement. Furthermore, all models are evaluated using the same BF16 precision and a fixed batch size of one to reflect real-world deployment scenarios accurately. Further details can be found in the {\color{blue} \bf APPENDIX A.2}.

\noindent {\bf Evaluation Metric}
These benchmarks include the MME-RealWorld~\cite{zhang2024mme} series, HR-Bench~\cite{wang2025divide}, and V\textsuperscript{*}~\cite{wu2024v}. We report results for different splits of each benchmark. For example, for the MME-RealWorld series, we report perception and reasoning accuracy separately. For HR bench, we report Fine-grained Single-instance Perception (FSP) and Fine-grained Cross-instance Perception (FCP) separately. For V\textsuperscript{*}, we report recognition and spatial relationship reasoning performance. Beyond these, we evaluate on Q-Test, JigSaw, and Relative Reflectance. These tasks are drawn from BLINK~\cite{fu2024blink}, a benchmark of expert-annotated, perception-heavy tasks designed for MLLMs.

\subsection{Main Result}

\noindent {\bf Comparison with the Baseline}
As illustrated in Table~\ref{tab:tab_main_results}, our EVA method demonstrates substantial improvements across multiple benchmarks. This performance gain is particularly pronounced on the MME-Real-Lite metric, where EVA achieves an improvement of $\bf 12.9\%$ over the baseline model Qwen2.5-VL-7B. 
Furthermore, it is noteworthy that our approach outperforms the larger Qwen2.5-VL-32B model on several key metrics, including FSP and overall scores in both HRbench-4K and HRbench-8K, as well as perception and reasoning tasks in MME-Real-Lite. Specifically, on MME-Real-Lite, EVA achieves a higher overall accuracy of $49.8$ compared to $46.2$ for Qwen2.5-VL-32B, despite the latter having four times the parameter count. 

\noindent {\bf Comparison with State-of-the-art Models}
We also conducted a comparative analysis against current state-of-the-art MLLMs, including both open-source and closed-source systems. For the open-source tool-using models, we selected the recent DeepEyes~\cite{zheng2025deepeyes} model for our comparison. As presented in Table~\ref{tab:tab_main_results}, our method outperforms these models across multiple benchmarks, particularly on tasks that require fine-grained visual perception and reasoning under challenging conditions. On the HRbench-4K benchmark, EVA achieves an overall accuracy of $74.0$ compared to $73.2$ for DeepEyes . These results highlight the superior performance of our approach in handling complex real-world visual scenarios.
Moreover, our approach demonstrates competitive performance compared to the closed-source GPT-4o model. EVA achieves comparable or better results on several key metrics, particularly in perception-heavy subtasks. For instance, on HRBench-8K, EVA reaches $69.5$ in perception, outperforming GPT-4o's $59.6$ in the same category.

\subsection{Ablation Study}
\begin{table}[]
\centering
\caption{{\bf Performance Comparison of EVA's Components.} For all open-source models, the best performance for each metric is {\bf bolded}, and the second best is {\ul underlined}. 
}
\label{tab:tab_aba1}
\scalebox{1.0}{

\scalebox{.65}{
\begin{tabular}{lcccccc}
\toprule
\multirow{2}{*}{Model}     & \multicolumn{3}{c}{HRbench-4K}                & \multicolumn{3}{c}{V*}                        \\
                           & FSP           & FCP           & Overall       & Attribute     & Spatial       & Overall       \\ \midrule 
Baseline                   & 85.2          & 52.2          & 68.8          & 78.2          & 73.6          & 76.4          \\
+ learning to generate     & 76.5          & \textbf{64.5} & 71.7          & 78.3          & {\ul 80.3}    & 79.1          \\
+ learning to use          & \textbf{90.7} & 55            & {\ul 72.9}    & {\ul 80.0}    & 78.9          & {\ul 79.5}    \\
\rowcolor{mycustomcolor} + learning to explore (EVA) & {\ul 90.5}    & {\ul 56.8}    & \textbf{73.7} & \textbf{80.0} & \textbf{80.3} & \textbf{80.2} \\ \bottomrule
\end{tabular}}
}
\end{table}

\begin{table}[t]
    \centering\setlength{\tabcolsep}{4.5pt}
\caption{{\bf Comparison of training efficiency.} EVA dramatically reduces the training time and computational resource requirements compared to DeepEyes, despite processing a significantly larger amount of SFT data.
}
\label{tab:tab_aba_training}
    \scalebox{.75}{
\begin{tabular}{lccccccc} 
\toprule
\multirow{2}{*}{Method} & \multicolumn{2}{c}{Hardware} & \multicolumn{2}{c}{Data Size} & \multirow{2}{*}{Rollout Size} & \multicolumn{2}{c}{Time ($\downarrow$)} \\ 
\cmidrule(lr){2-3} \cmidrule(lr){4-5} \cmidrule(lr){7-8} 
& SFT & RL & SFT & RL & & SFT & RL \\ 
\midrule
DeepEyes & - & 4 Nodes & - & 47K & 16 & - & $\sim$50h \\
\rowcolor{mycustomcolor}
EVA (Ours) & 4 Nodes & \textbf{1 Node} & 230K & 4K & $8$ & 20h & \textbf{5h} \\ 
\bottomrule
\end{tabular}
    }
\end{table}
Further experiments, including a comparison with bounding box methods, experiments on a 3B model, prompts and more visualization results, can be found in the {\color{blue} \bf APPENDIX C.1, C.2, C.3, C.4}.

\begin{figure*}[t]
  \centering
  \begin{subfigure}{0.45\linewidth}
    \includegraphics[width=1\linewidth]{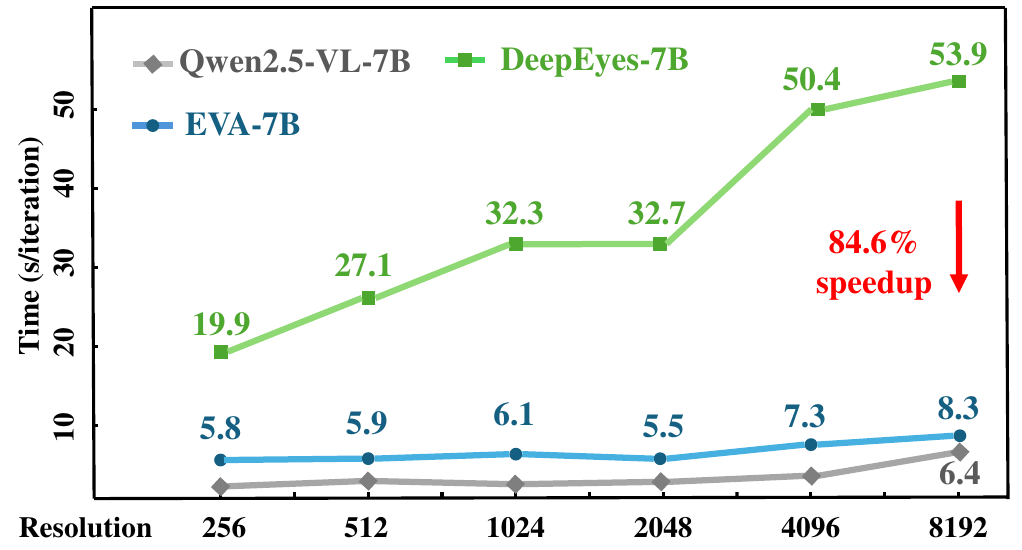}
    \label{fig:cifar-b0-10}
  \end{subfigure}
  \begin{subfigure}{0.45\linewidth}
    \includegraphics[width=1\linewidth]{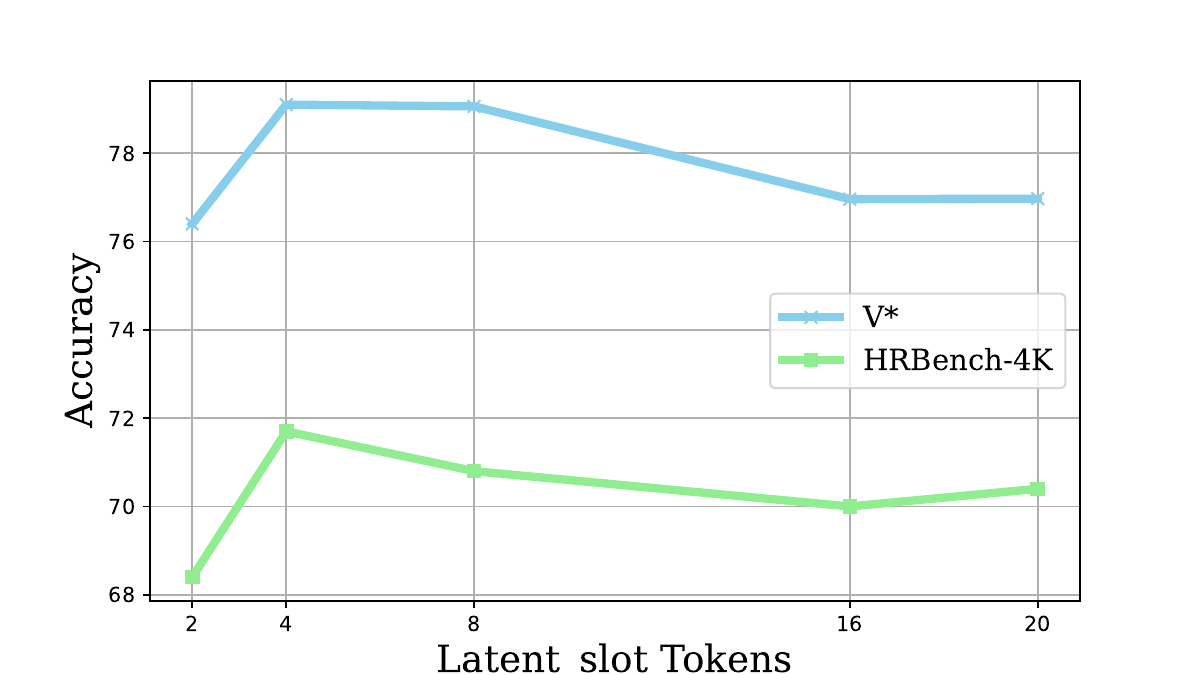}
    \label{fig:cifar-b50-10}
  \end{subfigure}
  \hfill
  \caption{(Left) Inference Latency Comparison across Different Image Resolutions. (Right) Effect of token length per Latent\_slot (\eg, learning to generate phase).}
  \label{fig:compare}
\end{figure*}
\begin{figure*}[t]
	\centering
        \includegraphics[width=1.0\textwidth]{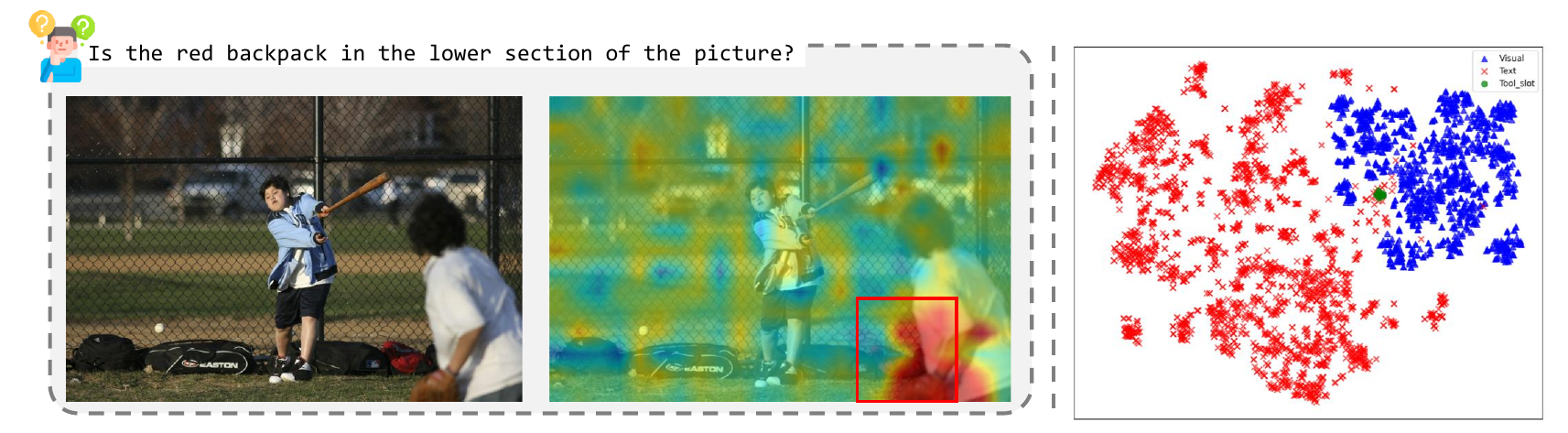}
        \caption{\textbf{Qualitative analysis of EVA framework.} {\color{red} \bf $\times$} represents text tokens, {\color{blue} \bf $\blacktriangle$} represents visual tokens, and {\color{ForestGreen} \bf $\bullet$} represents latent tokens.}
	\label{fig:fig_suppl_aba_visual}
\end{figure*}

\subsubsection{Ablation Study of Components in EVA.}
We conduct an ablation study to evaluate the contribution of each component in our EVA framework, as summarized in Table~\ref{tab:tab_aba1}. Starting from the baseline model Qwen2.5-VL-7B, we incrementally integrate the key stages of our training methodology—learning to generate, learning to use, and learning to explore—to assess their individual impact on performance.
Integrating the first stage increases the overall accuracy on HRbench-4K from 68.8 to 71.7, demonstrating that the ability to localize critical visual content significantly enhances perception performance.
Building upon this, the second stage leads to a further improvement in overall performance, rising to 72.9.
Finally, the third stage completes the full EVA framework and results in the best performance across all metrics: an overall accuracy of 74.0 on HRbench-4K and 80.2 on V\textsuperscript{*}.

\subsubsection{Efficiency Comparison}
Based on the efficiency analysis presented in Figure~\ref{fig:compare} (Left), we conduct a comparison of EVA-7B method against the tool-augmented DeepEyes-7B model, across different image resolutions. To ensure robust and stable measurements, the execution time reported for each model at each resolution is the average result of 50 independent inference runs. As shown in the figure, the inference time per iteration increases significantly with resolution for DeepEyes-7B, rising from $19.9$ seconds at $256 \times 256$ to $53.9$ seconds at $8192 \times 8192$. In contrast, EVA-7B maintains a nearly constant inference time of approximately $5.5$ to $8.3$ seconds across all resolutions, demonstrating its high computational efficiency and robustness to input scale.
At the highest resolution of $8192 \times 8192$, EVA achieves a remarkable $84.6\%$ speedup over DeepEyes, reducing the processing time from $53.9$ seconds to just $8.3$ seconds. 
These results collectively demonstrate that EVA not only matches or exceeds the accuracy of existing tool-augmented models but also achieves significantly higher inference efficiency, making it a more practical and scalable solution for complex visual reasoning tasks.
Table~\ref{tab:tab_aba_training} presents a comparison of training durations between EVA and DeepEyes. Although we utilize 230K samples for the two-stage SFT, EVA maintains a clear advantage in overall training time, thanks to the efficiency of using a smaller set of RL samples. These results collectively demonstrate the value of EVA for achieving high-efficiency inference.

\subsubsection{Ablation on the Token Length per \texttt{Latent\_slot}.}
Figure~\ref{fig:compare} (Right) shows the effect of the number of Latent\_slot tokens. 
As the number of tokens continues to increase, a slight decline in model performance is observed, suggesting that increasing the quantity of latent\_slot tokens beyond a certain point yields no additional benefits and instead compromises the stability of the optimization process.

\subsubsection{Latent Behavior Analysis.}
We utilize t-SNE to project pure textual tokens, visual embeddings, and the corresponding Latent\_slot representations into a shared low-dimensional space (\eg, dimension 2, perplexity 30).  
Figure\ref{fig:fig_suppl_aba_visual} (right) visualization reveals that the latent Latent\_slot tokens form a distinct cluster between textual and visual embeddings, demonstrating their role as a unified, modality-independent cross-modal representation.
Visualizing the EVA reasoning process, Figure\ref{fig:fig_suppl_aba_visual} (left)  shows EVA maintaining stable attention on task-relevant areas.

\subsubsection{Ablation on More Benchmarks}
\begin{table}[]
\centering
\caption{{\bf Performance Comparison on vision-centric tasks}. The result ($^{\dagger}$) is collected from \cite{lvr}.
{\color[HTML]{656565} \textit{Gray-colored}} font indicates improvement over the baseline Qwen2.5-VL-7B.
}
\label{tab:tab_aba4}
\scalebox{1.0}{
\begin{tabular}{lccc}
\toprule
Method                                      & IQ-Test                                & JigSaw                                 & Relative Reflect                    \\ \midrule
GPT-4o                                      & 30.0                                   & 58.0                                   & 38.8                                \\
Qwen2.5-VL-7B                               & 26.0                                   & 52.0                                   & 38.8                                \\
LVR-7B$^{\dagger}$                                         & 27.3                                   & 52.7                                   & 41.8                                \\
EVA-VL-7B                                         & 30.0                                   & 66.7                                   & 39.6                                \\
 \rowcolor{mycustomcolor}  {\color[HTML]{656565} \textit{Improvement}} & {\color[HTML]{656565} \textit{15.4\%}} & {\color[HTML]{656565} \textit{28.3\%}} & {\color[HTML]{656565} \textit{2\%}} \\ \bottomrule
\end{tabular}
}
\end{table}

In Table~\ref{tab:tab_aba4}, we conduct an ablation-style analysis to evaluate the effectiveness of our EVA-VL-7B model on vision-centric tasks. Compared to the baseline Qwen2.5-VL-7B, EVA-VL-7B achieves consistent improvements across all three benchmarks: IQ-Test ($+15.4\%$), JigSaw ($+28.3\%$), and Relative Reflect ($+2\%$). 
\section{Conclusion}
Tool-invocation methods based on discrete outputs introduce significant inference latency. To address these constraints, we introduced EVA, a novel framework that internalizes tool functionalities as latent intrinsic representations, named Tool\_slots, thereby eliminating the need for explicit command generation. We identified that the joint optimization of these latent representations with discrete tokens causes significant policy deviation, particularly in the "transition window" following the tool slot. We therefore developed the D-GSPO algorithm to resolve this core instability by decoupling the optimization of latent and discrete components. Our extensive experiments across multiple benchmarks confirm that EVA achieves substantial performance improvements and enhances inference efficiency. 

\newpage

{
    \small
    \bibliographystyle{ieeenat_fullname}
    \bibliography{main}
}
\newpage
\newpage
\appendix

\section{Details}
\label{sec:suppl_1}

\subsection{Dataset Construction}
\label{sec:suppl_1_1}

\begin{table*}[]
\centering
\caption{Statistics of the EVA-SFT-230K dataset. 
}
\label{tab:tab_aba_data}
\scalebox{0.8}{

\begin{tabular}{llll}
\hline
Source     & Domain                       & Visual Operation Type                        & Samples \\ \hline
Vsiual-CoT & Real-world, documents, chart & Cropping, drawing bounding box               & 0.2K    \\
Thyme      & Real-world, chart, OCR      & Cropping, rotation, low-contrast enhancement & 227.7K  \\
DeepEyes    & Real-world, chart            & Cropping                                     & 2K      \\
ReFocus    & Chart                        & Drawing bounding box, hightlighting          & 0.1K    \\ \hline
\end{tabular}
}
\end{table*}

EVA training data consists of two parts.: EVA-SFT-230K and EVA-RL-4K. For the supervised fine-tuning phase, we construct the EVA-SFT dataset, comprising approximately 230,000 samples integrated from Vision-CoT~\cite{shao2024visual}, ReFocus~\cite{fu2025refocus}, Thyme~\cite{zhang2025thyme}, and DeepEyes~\cite{zheng2025deepeyes}. 
While most samples in the aforementioned datasets provide post-tool-call visual content, a small portion lacks these corresponding images. For such cases, we prompt Qwen3 model to generate the necessary training data.
A critical feature of this subset is the variable number of Latent\_slots within each sample, which supports adaptive reasoning depths, ranging from 1 to 6 per sample. For the subsequent reinforcement learning stage, we compile the EVA-RL dataset, consisting of roughly 4,000 samples derived from DeepEyes~\cite{zheng2025deepeyes} and ThinkLite~\cite{wang2025sota}.
Table~\ref{tab:tab_aba_data} presents the tasks included in each dataset, the types of visual operations, and their corresponding proportions.

\subsection{Experimental Settings}
\label{sec:suppl_1_2}

\begin{table}[]
\centering
\caption{{\bf The setting of efficiency comparison}. 
}
\label{tab:tab_aba_hyper}
\scalebox{0.9}{

\begin{tabular}{lc}
\hline
Setting             & Value \\ \hline
GPU                 & H800  \\
memory              & 140G  \\
max response length & 4096  \\
test batch size     & 1     \\
top\_k              & 0.001 \\
top\_p              & 1     \\
temperature         & 0.01  \\
max pixels          & -     \\ \hline
\end{tabular}
}
\end{table}

Table~\ref{tab:tab_aba_hyper} details the environmental configurations for the inference efficiency comparison, with all experiments conducted under identical settings.

\section{Policy Deviation}
\label{sec:suppl_2}
\begin{figure}[t]
	\centering
        \includegraphics[width=0.45\textwidth]{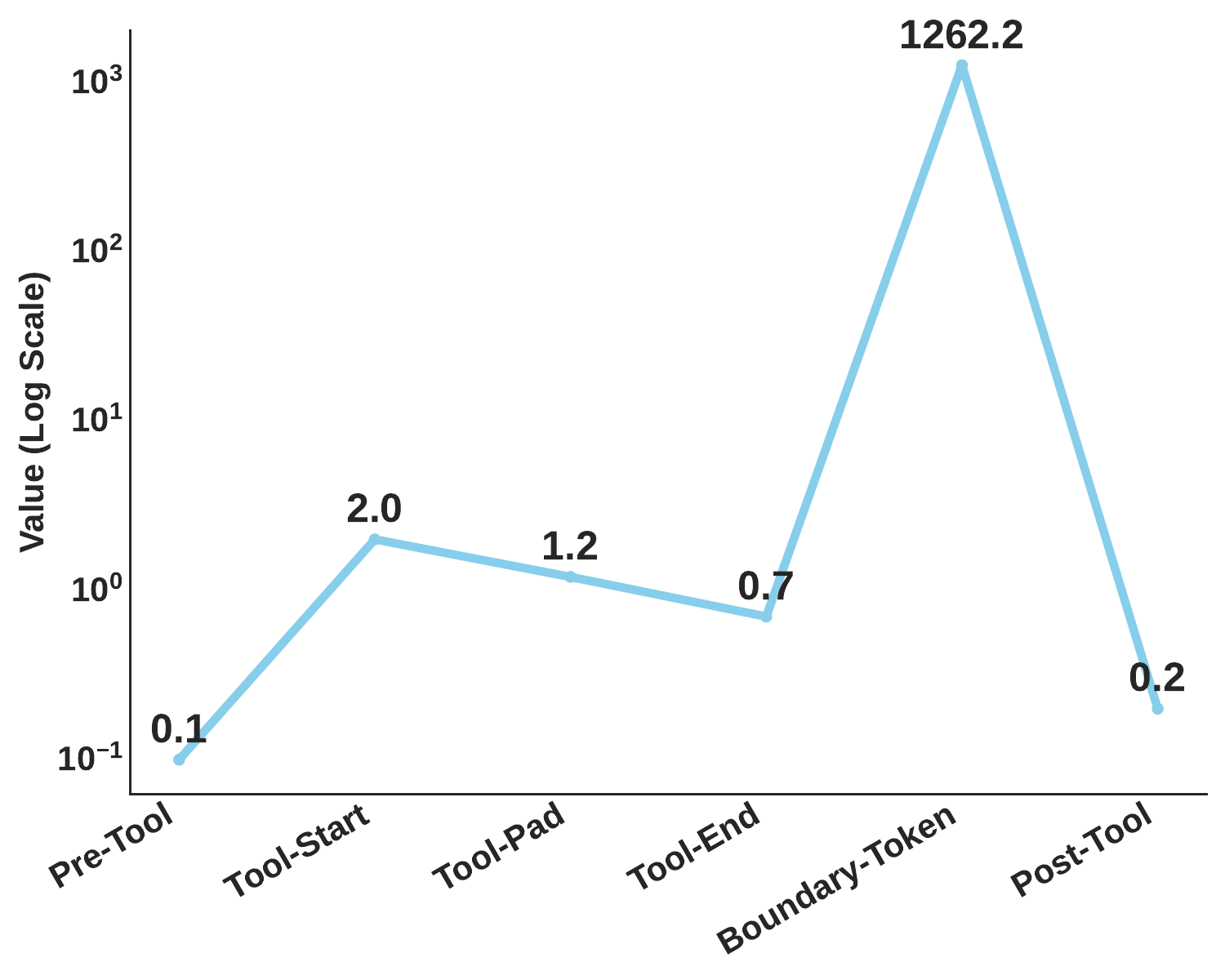}
        \caption{\textbf{Comparison of policy deviation across different positions in the output sequence.} }
	\label{fig:fig_suppl_aba3}
	\vspace{-1.5mm}
\end{figure}

During the initial reinforcement learning optimization, we observe a phenomenon where the model gradients exhibit explosive growth during training. This instability persists despite the application of various KL divergence strategies, different KL application positions, and the incorporation of auxiliary perception losses.
To investigate the root cause of this phenomenon, we analyze the policy deviation across different positions within the generated sequence. We visualize the policy deviation for general text tokens, the Latent\_slots, and the transition window (defined as the first five tokens immediately following a Latent\_slot), as shown in Figure~\ref{fig:fig_suppl_aba3}. The analysis reveals that the policy deviation within the transition window is significantly higher than in other regions. This discrepancy suggests that during the RL phase, the model suffers from catastrophic forgetting regarding the knowledge acquired during the Supervised Fine-Tuning (SFT) phase. Specifically, the model loses the ability to effectively utilize the Latent\_slot to guide subsequent reasoning steps.
To address this issue, we propose the D-GSPO algorithm in the main text. We also visualize the gradient trajectory after applying D-GSPO, which demonstrates that the gradients become significantly more stable compared to the original optimization process.

\begin{table}[]
\centering
\caption{{\bf Ablation study on different methods}.
}
\label{tab:tab_supple_aba2}
\scalebox{1.0}{
\begin{tabular}{lccc}
\toprule
\multirow{2}{*}{Method} & \multicolumn{3}{c}{V*}                        \\ \cline{2-4} 
                        & Attribute     & Spatial       & Overall       \\ \midrule
Qwen2.5-VL-7B           & 78.2          & 73.6          & 76.4          \\
Qwen2.5-VL-7B-box       & 77.4          & 79.0          & 78.0          \\
EVA-VL-7B-KL               & 80.0 & 79.0 & 79.6 \\ 
\rowcolor{mycustomcolor} EVA-VL-7B (ours)              & \textbf{80.0} & \textbf{80.3} & \textbf{80.2} \\ 
\bottomrule
\end{tabular}
}
\end{table}

\begin{table*}[]
\centering
\caption{{\bf Performance comparison on various tasks.}
}
\label{tab:tab_supple_aba1}
\scalebox{1.0}{
\begin{tabular}{lcccccc}
\toprule
\multirow{2}{*}{Method} & \multicolumn{3}{c}{V*}                        & \multicolumn{3}{c}{HRBench-4K}                \\ \cline{2-7} 
                        & Attribute     & Spatial       & Overall       & FSP           & FCP           & Overall       \\ \midrule
Qwen2.5-VL-3B           & \textbf{80.8} & 67.1          & 75.3          & 82.0          & 49.7          & 65.8          \\
\rowcolor{mycustomcolor} EVA-VL-3B               & 77.4          & \textbf{79.0} & \textbf{78.0} & \textbf{84.0} & \textbf{54.7} & \textbf{69.4} \\ \bottomrule
\end{tabular}
}
\end{table*}

\begin{figure*}[t]
	\centering
        \includegraphics[width=0.98\textwidth]{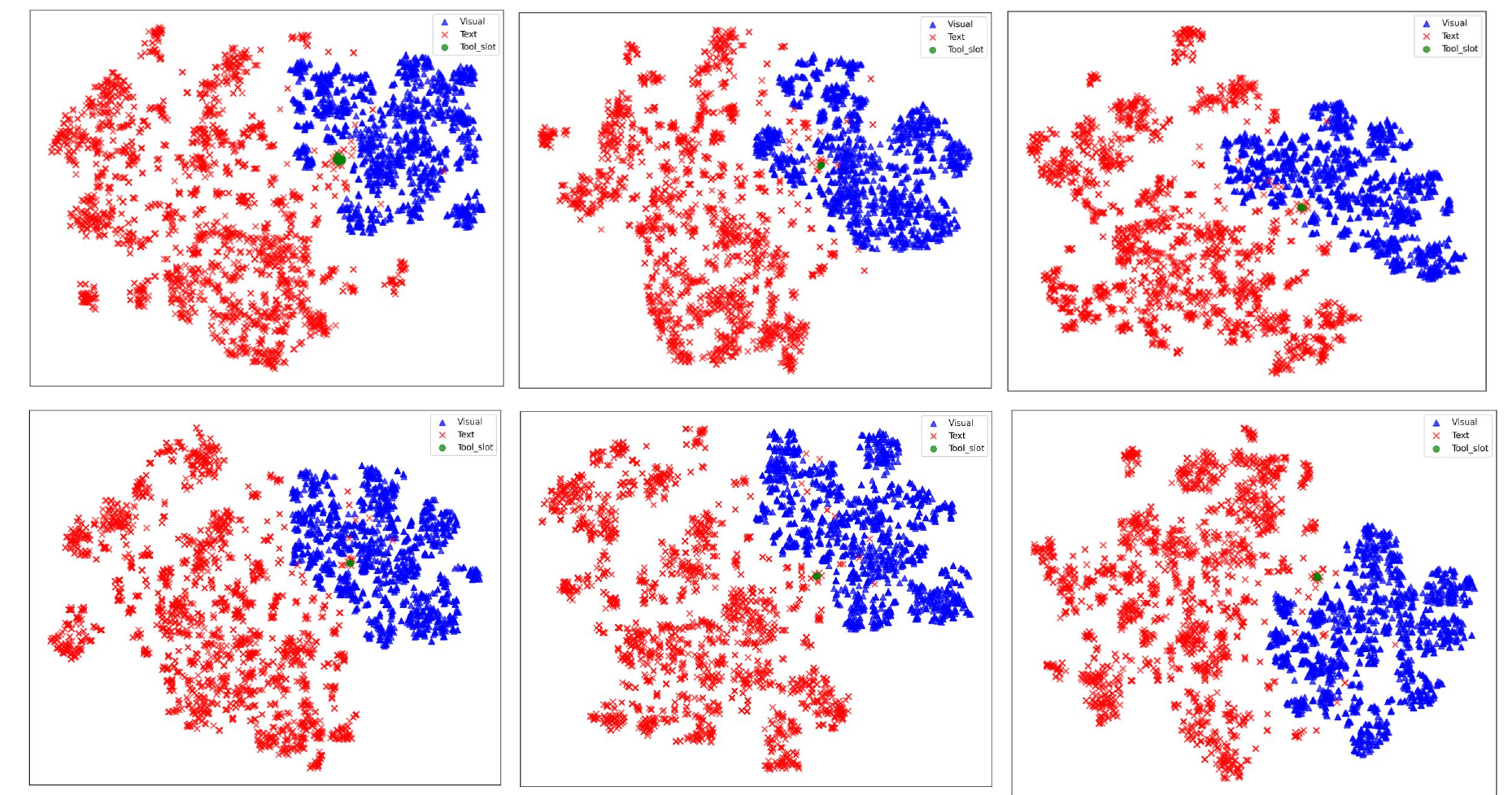}
        \caption{\textbf{Visualization of Tool\_slot embeddings}. {\color{red} \bf $\times$} represents text tokens, {\color{blue} \bf $\blacktriangle$} represents visual tokens, and {\color{ForestGreen} \bf $\bullet$} represents latent tokens.}
	\label{fig:fig_supple_aba1}
	\vspace{-1.5mm}
\end{figure*}
\begin{figure*}[t]
	\centering
        \includegraphics[width=1.0\textwidth]{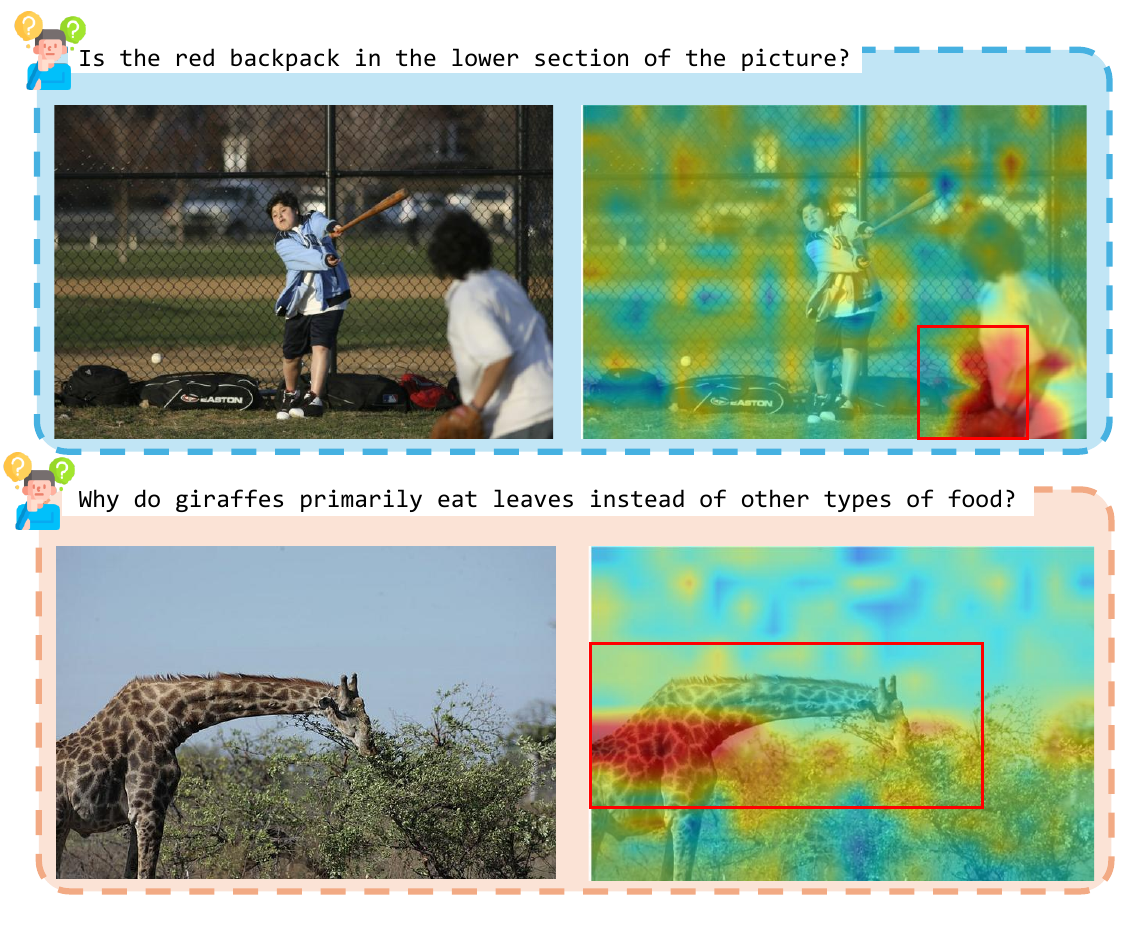}
        \caption{\textbf{Qualitative analysis of EVA framework.} }
	\label{fig:fig_visual}
	\vspace{-1.5mm}
\end{figure*}

\section{Additional Ablation Studies}
\label{sec:suppl_3}

\subsection{Comparison with Bounding Box and Vanilla-GSPO Methods}
To investigate the impact of bounding box prediction on multimodal reasoning performance, we conduct an ablation study comparing three variants: the baseline Qwen2.5-VL-7B model, its extension with explicit box output (Qwen2.5-VL-7B-box), and our proposed EVA-VL-7B framework. As shown in Table~\ref{tab:tab_supple_aba2}, 
our proposed EVA-VL-7B achieves remarkable performance. Crucially, it significantly outperforms the explicit box output variant, validating the superiority of our approach.
We provide comprehensive ablation studies of D-GSPO in the main text, while Table~\ref{tab:tab_supple_aba2} details the ablations on the strong KL loss. Notably, applying distinct KL constraints to transition window tokens and other tokens also yields performance gains.

\subsection{Experiments on a 3B Model}
To evaluate the scalability of our approach, we replicate the experiments using a smaller 3B parameter model. As shown in Table~\ref{tab:tab_supple_aba1}, our method consistently achieves performance gains over the baseline counterpart. This finding confirms that the efficacy of the EVA framework generalizes well to smaller model architectures.

\begin{table}[]
\centering
\caption{{\bf Performance Comparison on the MME-Realworld}. 
{\color[HTML]{656565} \textit{Gray-colored}} font indicates improvement over the baseline Qwen2.5-VL-7B.
}
\label{tab:tab_aba3}
\scalebox{0.75}{
\begin{tabular}{lccc}
\toprule
\multicolumn{4}{c}{Perception}                                   \\ \midrule
Model         & Monitoring & Autonomous Driving & Remote Sensing \\
Qwen2.5-VL-7B & 38.8       & 22.7               & 45.4           \\
EVA-VL-7B     & 43.3       & 33.8               & 48.6           \\
 \rowcolor{mycustomcolor}  {\color[HTML]{656565} \textit{Improvement}}   & {\color[HTML]{656565} \textit{11.6\%}}     & {\color[HTML]{656565} \textit{48.9\%}}             & {\color[HTML]{656565} \textit{7\%}}            \\ \midrule
\multicolumn{4}{c}{Reasoning}                                    \\ \midrule
Model         & Monitoring & Autonomous Driving & Remote Sensing \\
Qwen2.5-VL-7B & 26.1       & 24.3               & -              \\
EVA-VL-7B     & 32.1       & 30.1               & -              \\
 \rowcolor{mycustomcolor}  {\color[HTML]{656565} \textit{Improvement}}   & {\color[HTML]{656565} \textit{23.0\%}}     & {\color[HTML]{656565} \textit{23.9\%}}             & -              \\ \bottomrule
\end{tabular}
}
\end{table}

\subsection{More Benchmarks}
In Table~\ref{tab:tab_aba3}, we conduct a concise ablation analysis on the MME-Realworld benchmark, comparing EVA-VL-7B against Qwen2.5-VL-7B across perception and reasoning tasks. EVA achieves consistent gains: $+11.6\%$ in Monitoring, $+48.9\%$ in Autonomous Driving, and $+7\%$ in Remote Sensing. In reasoning, improvements of $+23.0\%$ and $+23.9\%$ are observed in Monitoring and Autonomous Driving, respectively. These results demonstrate that our method significantly enhances both visual understanding and logical inference in real-world scenarios, particularly in complex domains like autonomous driving where accurate scene interpretation is critical.

\subsection{Visualization}

\subsubsection{Visualization of Latent Embeddings}
Adopting the visualization methodology, we utilize t-SNE to project pure textual tokens, visual embeddings, and the corresponding Latent\_slot representations into a shared low-dimensional space (\eg, dimension 2, perplexity 30). As illustrated in Figure~\ref{fig:fig_supple_aba1}, the textual components are widely dispersed across the manifold, whereas the visual embeddings form a distinct and compact cluster. 
Notably, the representations of our trained Latent\_slots are positioned in a unified, modality-independent cross-modal representation.

\subsubsection{Visualization of Reasoning Perception}
Due to the inherent lack of interpretability of latent embeddings, the \textit{cropping} task offers a more intuitive platform for visualization compared to other tasks. In Figure~\ref{fig:fig_visual}, we provide more attention heatmaps for the cropping task, which reveal that the \texttt{latent\_slot} accurately focuses on task-relevant regions.

\subsubsection{Visualization of SFT data instances}
In Table~\ref{fig:fig_example2}, we present more comprehensive examples from the EVA-SFT-230K training dataset to further illustrate its diversity.

\subsection{Details of Latent\_slot}
During the SFT phase, to ensure the effectiveness of feature compression, we filter the dataset by removing samples where the feature dimension of the tool image is less than 20. 
We allocate a corresponding number of \texttt{Latent\_slot} placeholders based on the quantity of tool images present in each sample. To guarantee accurate alignment during the matching process with the ground truth, we use order constraints. 
Furthermore, in the Reinforcement Learning (RL) phase, we discard a small number of samples generated during the rollout process where the quantity of predicted \texttt{Latent\_slot} tokens does not match the number of reserved positions.

\subsection{EVA Generated Examples}
\begin{figure*}[t]
	\centering
        \includegraphics[width=1.0\textwidth]{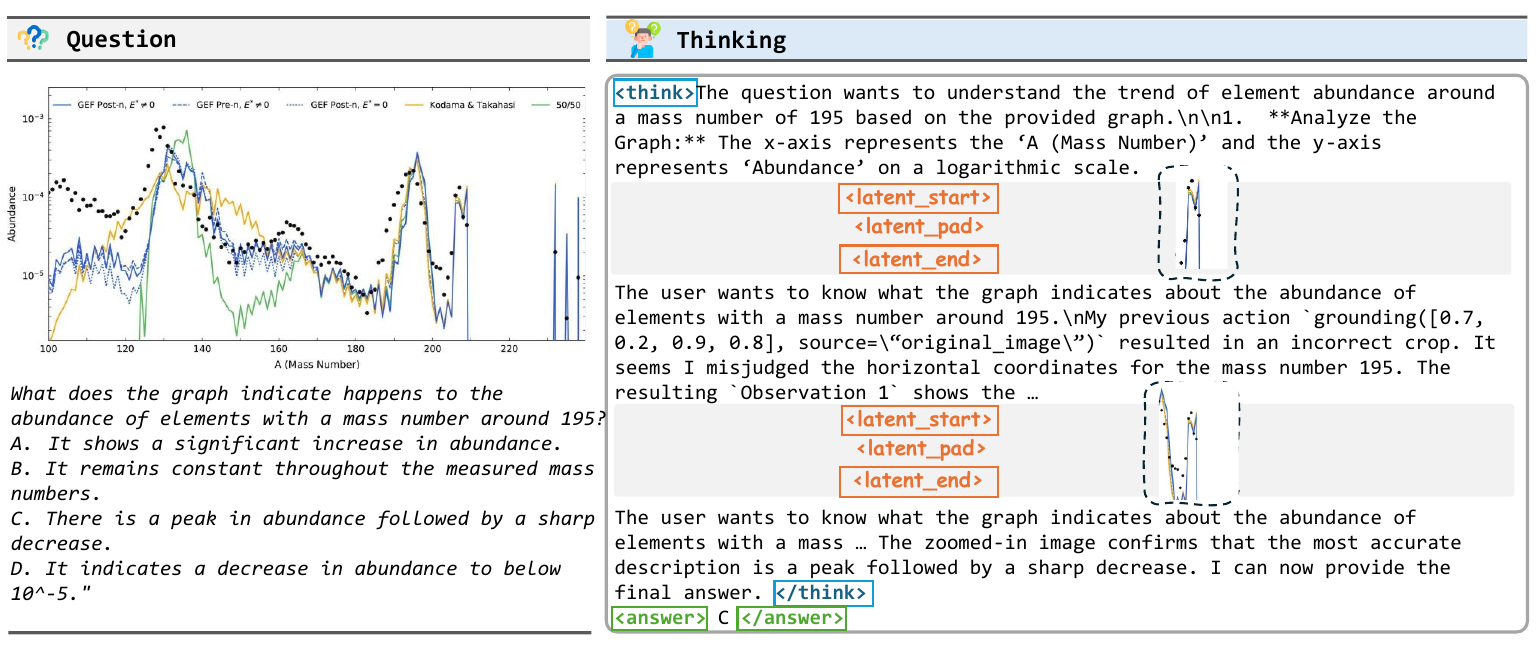}
        \caption{\textbf{Visualization of SFT data instances}}
	\label{fig:fig_example2}
	\vspace{-2.5mm}
\end{figure*}

\begin{figure*}[t]
	\centering
        \includegraphics[width=1.0\textwidth]{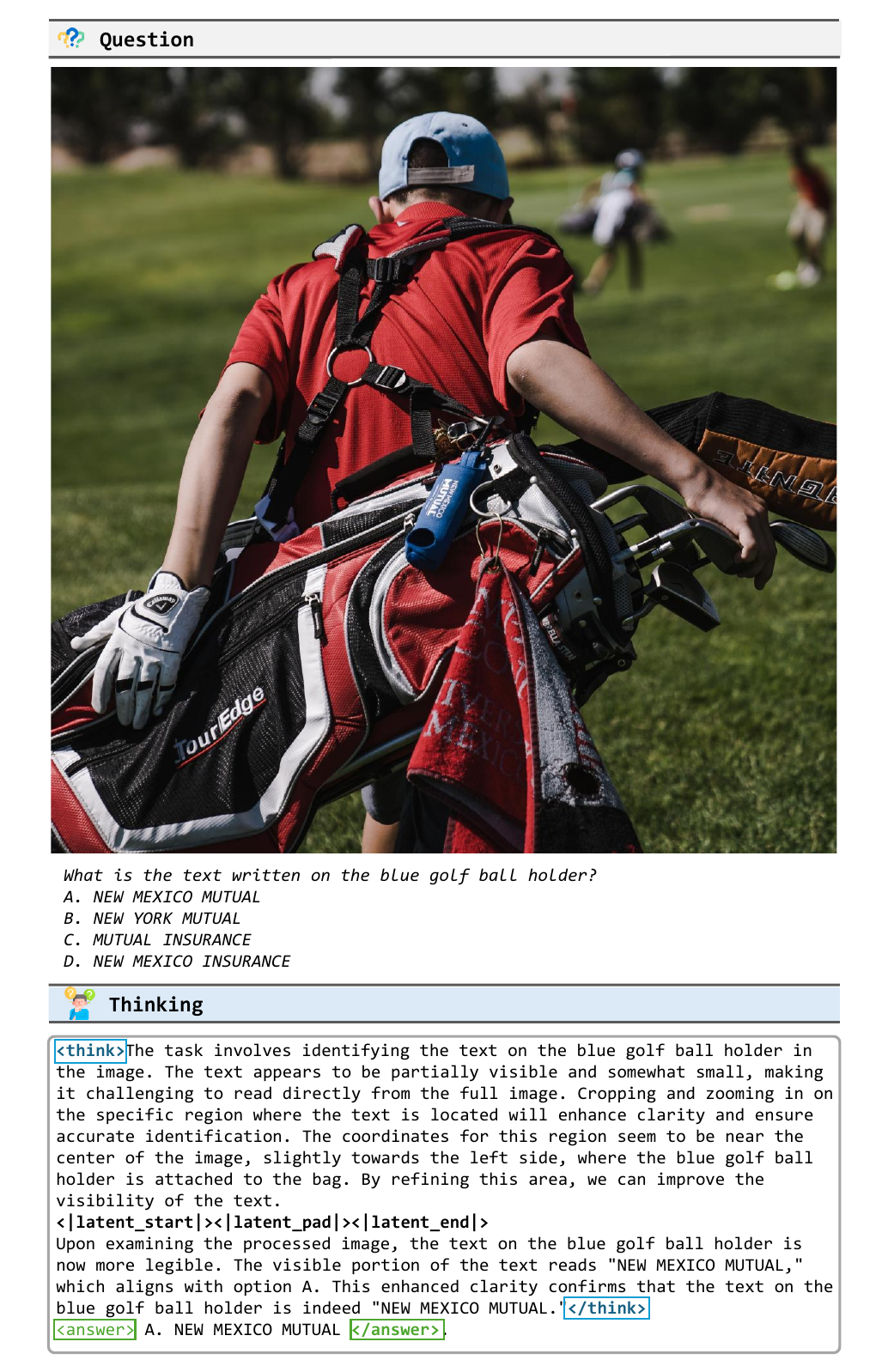}
        \caption{\textbf{Rotation Example}.}
	\label{fig:fig_output_1}
	\vspace{-2.5mm}
\end{figure*}

\begin{figure*}[t]
	\centering
        \includegraphics[width=0.9\textwidth]{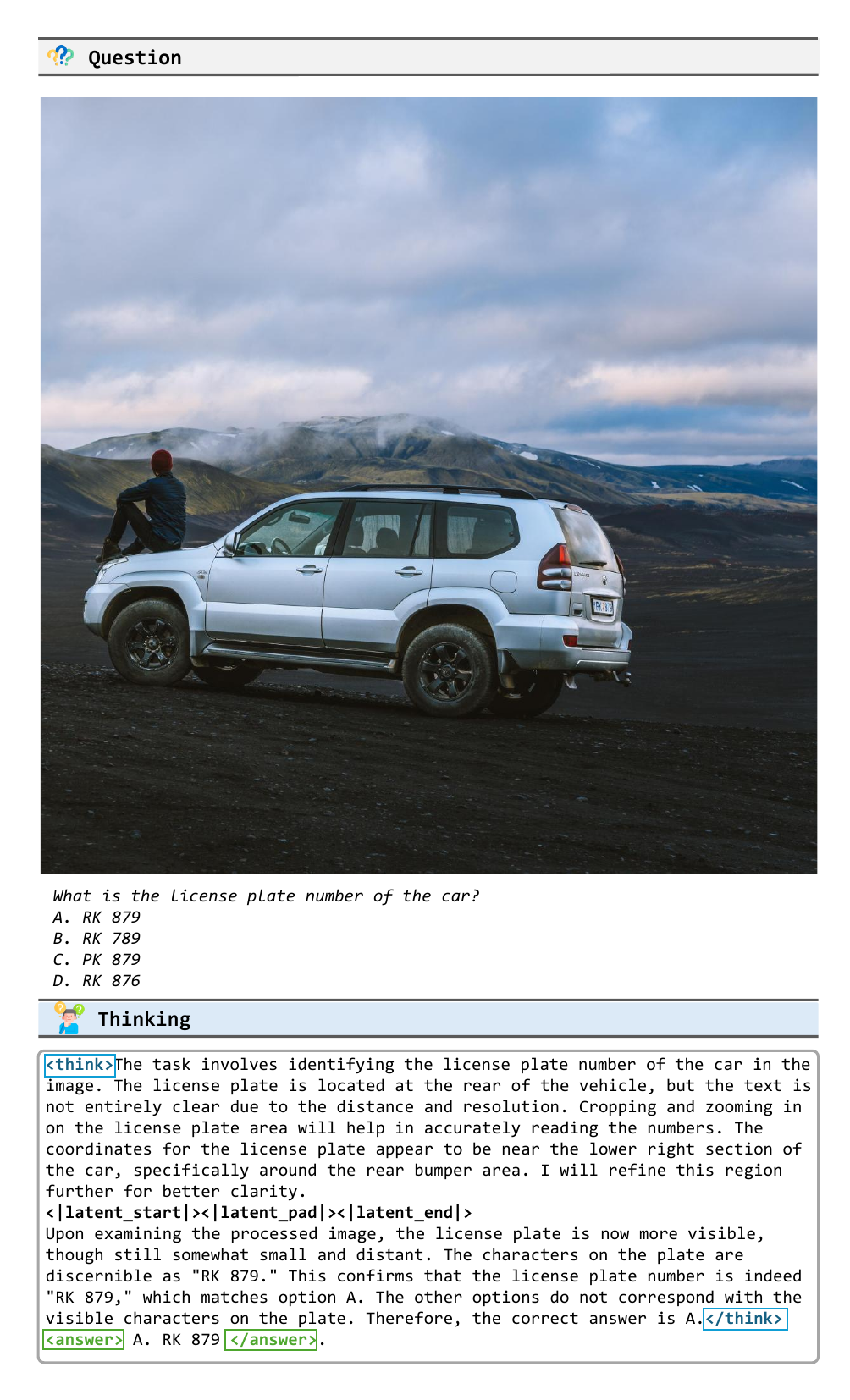}
        \caption{\textbf{Crop Example}.}
	\label{fig:fig_output_2}
	\vspace{-2.5mm}
\end{figure*}

\begin{figure*}[t]
	\centering
        \includegraphics[width=1.0\textwidth]{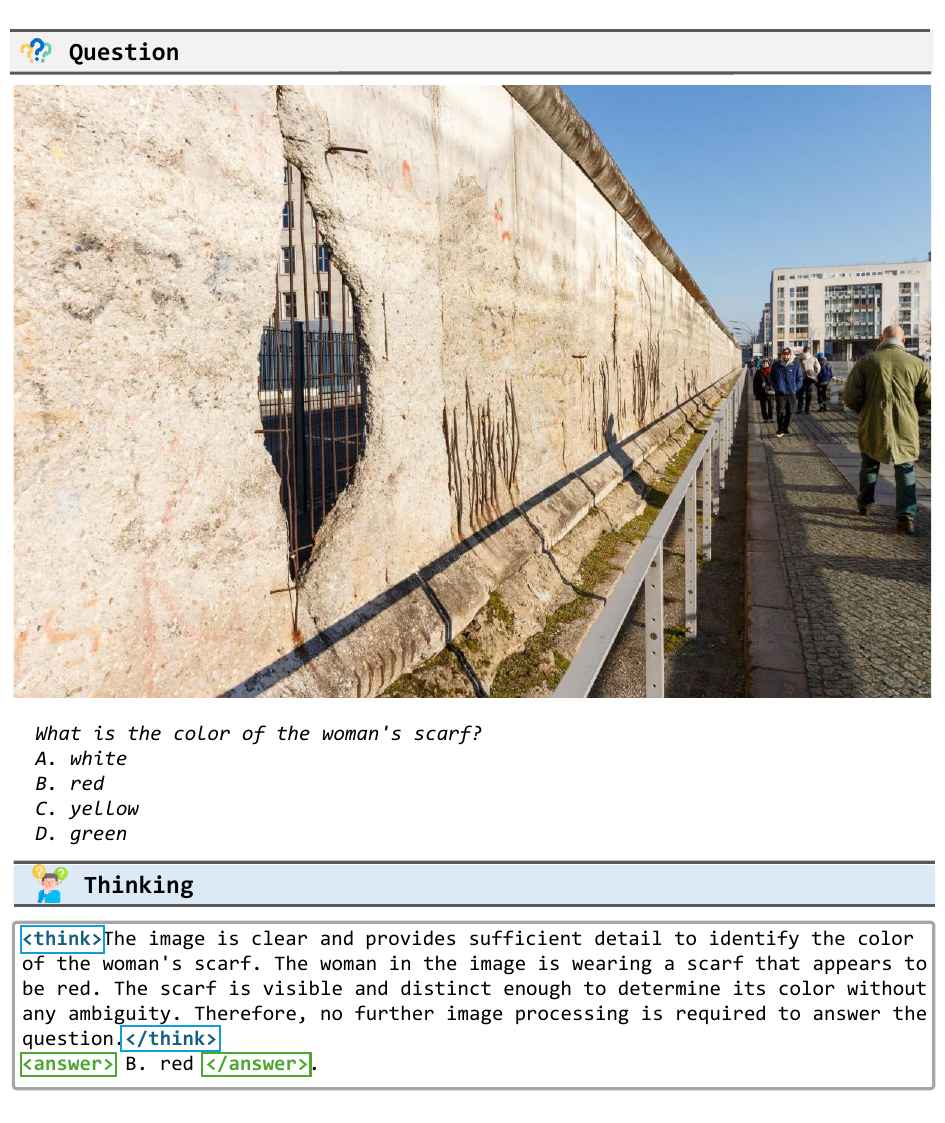}
        \caption{\textbf{Easy Example}}
	\label{fig:fig_output_3}
	\vspace{-2.5mm}
\end{figure*}

Figures~\ref{fig:fig_example2}, Figures~\ref{fig:fig_output_1}, Figures~\ref{fig:fig_output_2}, Figures~\ref{fig:fig_output_3} provide visualizations of scenarios involving low-resolution images with blurred details, simple problems where original images are clear, and content-flipped cases, respectively. It can be observed that while the EVA model answers directly for clear images, it adaptively generates corresponding image \texttt{latent\_slots} to assist reasoning when encountering low-quality or flipped content.

\subsection{Bad Cases}
\begin{figure*}[t]
	\centering
        \includegraphics[width=1.0\textwidth]{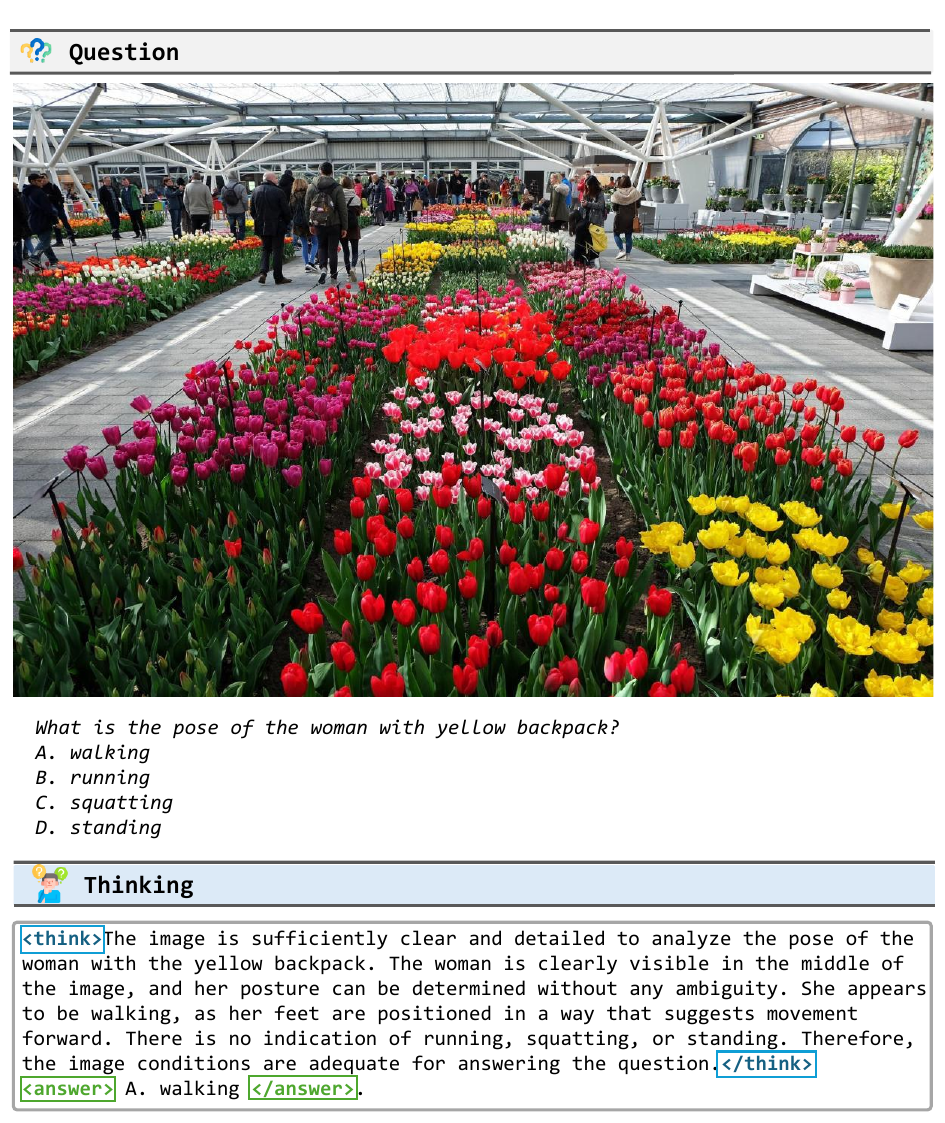}
        \caption{\textbf{Inaccurate cropping.}}
	\label{fig:fig_output_4}
	\vspace{-2.5mm}
\end{figure*}

Figure~\ref{fig:fig_output_4} illustrates examples where the attention focuses on incorrect locations. When the background elements are complex or the target object mentioned in the task appears in multiple places within the image, EVA still exhibits certain limitations.

\section{Prompts}
The prompts used for data construction in the SFT stage are provided below.

\section{Future Works}
We will conduct an in-depth investigation into the optimization of latent representations during reinforcement learning.

\section{Limitation}
We acknowledge that a disparity exists between our current data scale and that of state-of-the-art approaches. 
The optimization of latent representations through reinforcement learning offers a landscape for further exploration.
Consequently, our model does not currently exhibit a significant performance lead over tool-invocation methods. 
Despite this limitation, the inherent efficiency of the EVA paradigm suggests substantial potential for future exploration.

\section{Related Work}
\label{sec:suppl_0}
\subsection{Thinking about Images}
Multimodal Large Language Models (MLLMs)~\cite{jaech2024openai,chen2025transmamba,guo2025deepseek} have witnessed rapid advancements in recent years, leading to significant breakthroughs in visual understanding and cross-modal reasoning within the field of artificial intelligence.
The introduction of OpenAI o1~\cite{jaech2024openai} establishes the modern paradigm of inference time scaling methods, where native long-chain reasoning substantially boosts MLLMs’ reasoning capabilities. Subsequent open-source efforts have advanced this direction along two dominant axes: supervised fine-tuning (SFT)-based optimization and reinforcement learning (RL)-based optimization.
SFT-based approaches adapt pretrained MLLMs using curated or distilled multimodal reasoning corpora, enabling models to learn explicit step-by-step reasoning strategies~\cite{cot,Shi2024MathLLaVABM,Gao2023GLLaVASG}. Representative efforts such as LLaVA-CoT~\cite{LLaVACoT}, AtomThink~\cite{AtomThink}, and LLaMA-Berry~\cite{LLaMABerry} extend classical SFT pipelines by incorporating teacher-generated CoT traces and step-wise search strategies. 
The emergence of DeepSeek-R1~\cite{guo2025deepseek} then brought RL-based approaches to the forefront, with its verified rewards offering more reliable and robust supervisory signals. Building on this paradigm, open-source RL variants such as MMR1~\cite{MMR1}, Vision-R1~\cite{VisionR1}, R1-VL~\cite{R1VL} and MM-Eureka~\cite{meng2025mm} enhance self-verification through rule-based reward mechanisms. 
Specially, Vision-R1~\cite{huang2025vision} utilizes GRPO, coupled with hard format reward functions and cold-start initialization data, to improve emergent reasoning. 
Combining the strengths of both SFT and RL, systems like R1-OneVision~\cite{R1Onevision} and Mimo-VL~\cite{MiMoVL} employ multi-stage post-training to iteratively refine policy models while maintaining a balance between advanced reasoning ability and general-purpose performance.

\subsection{Thinking with Images}
Recent progress in tool-augmented large models~\cite{team2025kimi,geng2025webwatcher,mai2025agent,tao2025webshaper,xue2025simpletir} has introduced a complementary direction for enhancing reasoning by enabling models to interact with external computational environments. Early tool-series frameworks such as ReTool~\cite{Feng2025ReToolRL}, Sketchpad~\cite{Hu2024VisualSS}, C2-Evo~\cite{Chen2025C2EvoCM}, DeepEyes~\cite{zheng2025deepeyes}, Chain-of-Focus~\cite{zhang2025chain}, Pixer Reasoner~\cite{su2025pixel} focus on tightly integrating symbolic solvers, code execution engines, or structured verification modules into the reasoning loop, allowing models to offload algebraic manipulation, equation solving, or geometric computation to specialized tools. More recent tool-augmented architectures, including Active-o3~\cite{zhu2025active}, MGPO~\cite{huang2025high}, Thyme~\cite{zhang2025thyme} and Mini-o3~\cite{lai2025mini}, extend this paradigm by supporting richer tool ecosystems, dynamic tool selection, and multi-step tool–LLM interaction protocols. These models emphasize adaptive reasoning, where LLMs iteratively call external tools, interpret outputs, and refine their intermediate reasoning traces. Together, these tool-integrated methods form an emerging line of research that complements SFT and RL-based mathematical reasoning approaches by leveraging external computational modules to enhance correctness, reduce hallucination, and enable more faithful problem-solving pipelines.

\begin{figure*}[t]
	\centering
        \includegraphics[width=1.0\textwidth]{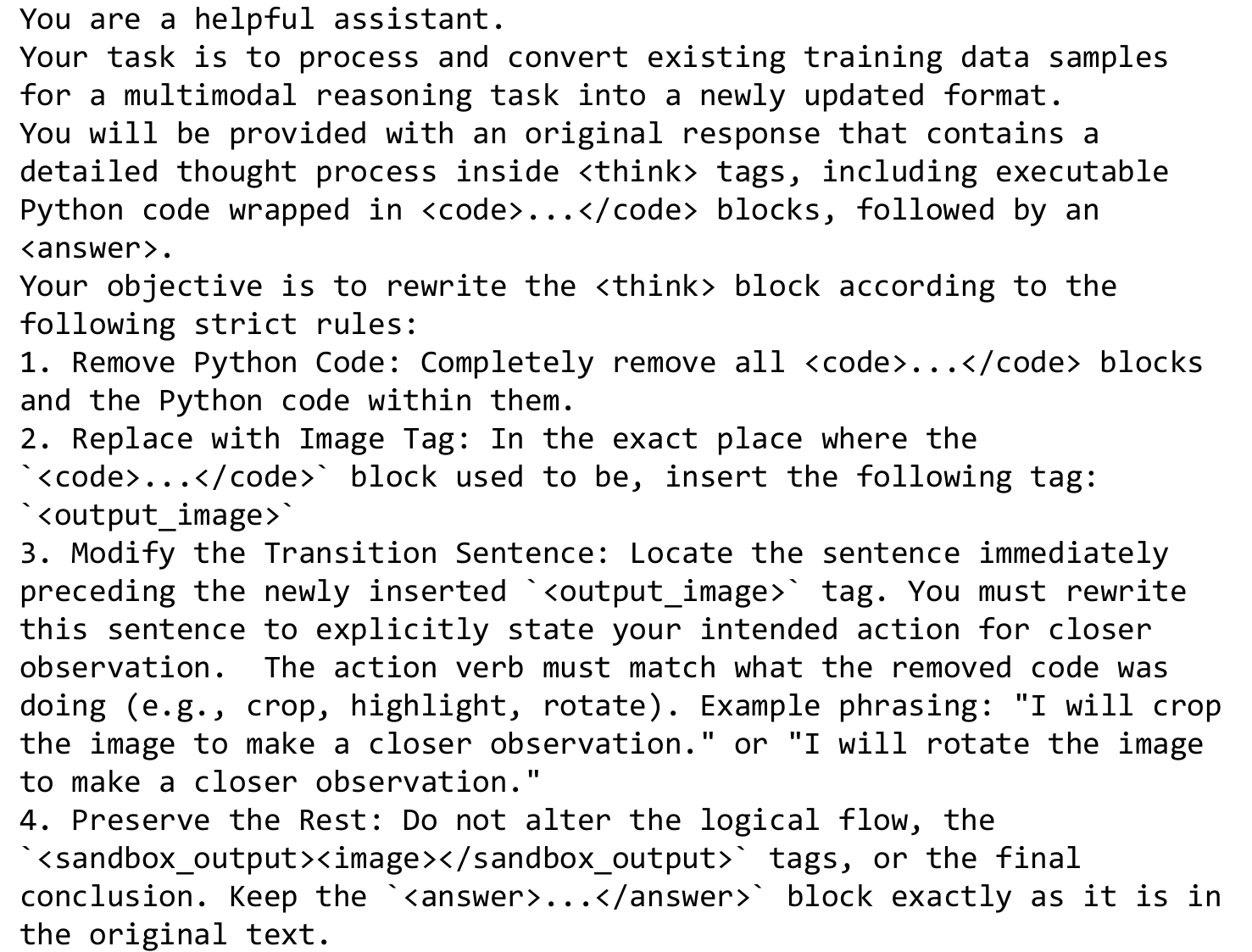}
	\label{fig:fig_prompt}
	\vspace{-2.5mm}
\end{figure*}
\begin{figure*}[t]
	\centering
        \includegraphics[width=1.0\textwidth]{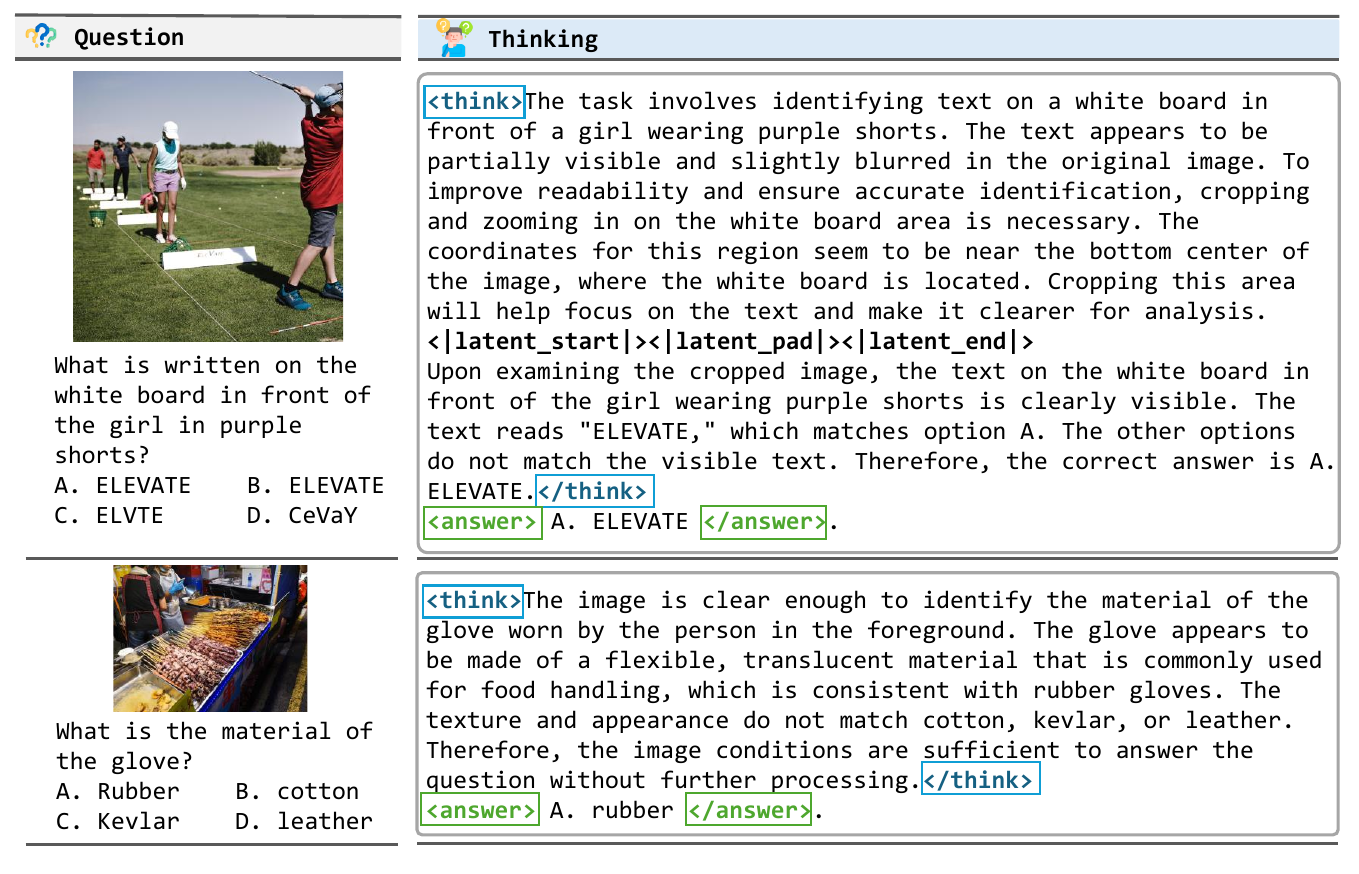}
        \caption{\textbf{Qualitative Examples of EVA.}}
	\label{fig:fig_aba_output}
	\vspace{-2.5mm}
\end{figure*}

\end{document}